\documentclass{article}

    \PassOptionsToPackage{numbers, compress}{natbib}
 \usepackage[preprint]{neurips_2026}

\usepackage[utf8]{inputenc} % allow utf-8 input
\usepackage[T1]{fontenc}    % use 8-bit T1 fonts
\usepackage{hyperref}       % hyperlinks
\usepackage{url}            % simple URL typesetting
\usepackage{booktabs}       % professional-quality tables
\usepackage{amsfonts}       % blackboard math symbols
\usepackage{nicefrac}       % compact symbols for 1/2, etc.
\usepackage{microtype}      % microtypography
\usepackage{xcolor}         % colors
\usepackage{graphicx}
\usepackage{eccvabbrv}
\usepackage{multirow} % Required for multirows
\usepackage{pifont}
\usepackage[misc]{ifsym}
\usepackage{amsmath}
\usepackage{caption}

\title{One-Shot Feed-Forward 360$^{\circ}$ Animatable Avatar \\via Inpainted UV-Space Gaussian Modeling
}

\author{
  \textbf{Shuling Zhao}
   \hspace{3cm}
  \textbf{Dan Xu}\textsuperscript{\Letter} \\ 
  \texttt{szhaoax@connect.ust.hk} 
  \hspace{1cm}
  \texttt{danxu@cse.ust.hk} \\
  The Hong Kong University of Science and Technology \\ 
  Zeekr Automobile R\&D Co., Ltd.
}

\begin{document}

\maketitle

\begin{abstract}
Building one-shot 3D animatable head avatars is an important yet challenging problem. Existing methods generally collapse under large camera pose variations, compromising the realism of 3D avatars. In this work, we propose a new framework to tackle the novel setting of one-shot 3D full-head animatable avatar reconstruction in a single forward pass via inpainted UV-space Gaussian modeling, enabling 360$^\circ$ rendering views and real-time animation. To facilitate efficient animation control, we model 3D head avatars with Gaussian primitives embedded on the surface of a parametric face model within the UV space, and project the input image features to the UV space, resulting in incomplete local UV feature maps. To inpaint the missing regions, we obtain knowledge of full-head geometry and textures from rich 3D full-head priors within a pretrained 3D generative adversarial network (GAN) for global full-head feature extraction and multi-view supervision. Specifically, to enhance the fidelity of 3D reconstruction during inpainting, we take advantage of the symmetric nature of the UV space and human faces to fuse incomplete yet detailed local UV feature maps with the extracted global full-head textures, resulting in inpainted UV Gaussian attribute maps for avatar modeling. Extensive experiments demonstrate \textcolor{black}{that our method is the first to achieve high-quality 3D full-head animatable avatar modeling, significantly improving side and back views while outperforming state-of-the-art animation approaches}, thereby improving the realism of 3D animatable avatars. 
% \textcolor{red}{
The project page is available at \href{https://shaelynz.github.io/fhavatar/}{https://shaelynz.github.io/fhavatar/}.
% }
% \keywords{3D Gaussian Splatting \and Animatable Head Avatar \and One-shot 3D Reconstruction} 
\end{abstract}

% \vspace{-8pt}
\section{Introduction}
\label{sec:intro}
\vspace{-2pt}

% \begin{center}
%     \centering
%     \captionsetup{type=figure}
%   \includegraphics[width=1\textwidth]{figures/teaser_refined.pdf}
%   \vspace{-18pt}
%   \caption{Given a single input image, our method reconstructs animatable 3D full-head Gaussian avatars in a single forward pass. These avatars provide consistent, high-fidelity 360° view synthesis and support real-time (246 FPS) animation.
% \end{center}

3D animatable head avatar creation aims to reconstruct 3D animatable heads from 2D inputs, with widespread applications such as teleconferencing, virtual reality (VR), and augmented reality (AR). Despite the success achieved in 3D head reconstruction from multi-view or monocular videos~\cite{qian2024gaussianavatars, zheng2023pointavatar, xiang2024flashavatar, zielonka2023instant}, the requirement for video-specific optimization makes such methods time-consuming. To improve efficiency, some approaches reconstruct 3D heads from a single image in a feed-forward manner~\cite{li2023generalizable, tran2024voodoo, ye2024realdportrait, deng2024portrait4d, deng2024portrait4d_eccv} using Neural Radiance Fields (NeRF)~\cite{mildenhall2020nerf}. 
%However, the slow rendering speed hinders real-time applications. 
Recently, 3D Gaussian Splatting (3DGS)~\cite{kerbl20233d} is adopted~\cite{chu2024generalizable, he2025lam, guo2025sega} for its impressive rendering quality and speed. 
%Nevertheless, these methods cannot maintain 3D consistency under large changes in camera pose.

Despite the advancement made in generalizable one-shot 3D animatable head avatar reconstruction, existing approaches generally fail when the rendering views significantly differ from the input camera pose. Trained on large-scale monocular face video datasets~\cite{xie2022vfhq, zhu2022celebv}, which predominantly capture frontal face motions, 
\textcolor{black}{these methods are fundamentally limited to near-frontal or partial views, and cannot handle the severe ambiguity of unseen regions (\eg, side/back head),} which harms the realism of the reconstructed 3D heads. 
%they can hardly generalize to large side views or even back views, 
% which harms the realism of the reconstructed 3D heads. 

To realize efficient avatar animation and $360^\circ$ view synthesis at the same time, we introduce a new framework for the novel setting of one-shot feed-forward 3D full-head animatable avatar creation via inpainted UV-space Gaussian modeling. To allow for animation control, we generate Gaussian primitives in the UV space of a parametric face model (FLAME~\cite{li2017learning}) so that Gaussians are rigged to the mesh surface and can move along with it during animation. Since the input image only provides appearance information from a single view, projecting its features to the UV space results in incomplete local UV feature maps. To obtain knowledge of full-head geometry and textures, inspired by the recent success of 3D GANs in full-head generation~\cite{an2023panohead, li2024spherehead}, we utilize a pretrained 3D GAN~\cite{an2023panohead} and its feed-forward inversion method~\cite{bilecen2024dual} to efficiently extract UV-space global full-head features from the input image and provide multi-view supervision during training.
% To obtain knowledge of full-head geometry and textures, inspired by the recent success of 3D generative adversarial networks (GANs) in full-head generation~\cite{an2023panohead, li2024spherehead}, we utilize a pretrained 3D GAN~\cite{an2023panohead} and its feed-forward inversion method~\cite{bilecen2024dual} to efficiently extract UV space global full-head features from the input image and provide multi-view supervision during training.
%To overcome the limited reconstruction fidelity of 3D GAN inversion, 
As 3D GAN inversion has limited reconstruction fidelity, we inpaint the incomplete but fine-grained local UV features with the \textcolor{black}{structured} global UV full-head feature. Specifically, to make full use of the input view, we leverage the symmetric nature of human faces and the UV space with a transformer-based feature fusion architecture. This approach allows tokens from the global UV map to query symmetric positions on the incomplete local UV feature maps for input appearances, \textcolor{black}{guiding the recovery of local details, which results} in inpainted UV features for Gaussian avatar modeling. To further improve the 3D visual quality of the reconstructed avatar, we propose a 3D total variation loss that encourages Gaussians to fully cover the whole avatar surface, alleviating hole-like artifacts.

%On the other hand, recent 3D generative adversarial networks (GANs) have shown promising capability in full-head generation~\cite{an2023panohead, li2024spherehead}. Trained on both frontal and back views of human heads, they can provide plausible 3D full-head appearances for input images through GAN inversion~\cite{roich2022pivotal, bilecen2024dual}. Motivated by their success, we introduce a novel framework for 3D full-head animatable Gaussian avatar reconstruction. 

%To enable avatar animation while reconstructing the geometry of the 3D full head, we generate Gaussians in the UV space of a parametric face model (FLAME~\cite{li2017learning}), binding them to the mesh surface. To obtain the invisible appearance of the input image, we utilize a pretrained 3D GAN~\cite{an2023panohead} and its feed-forward inversion method~\cite{bilecen2024dual} to generate the global head feature and project it to the UV space. 

Extensive experiments \textcolor{black}{confirm} that our \textcolor{black}{method} not only \textcolor{black}{is the first to reconstruct} 3D full-head avatars that can be viewed in 360$^\circ$, but also \textcolor{black}{surpasses previous methods in avatar animation performance}. %achieves state-of-the-art performance in avatar animation.
Our contributions are summarized as follows:
\begin{itemize}%[leftmargin=*]
    % \item To the best of our knowledge, this paper is the first to create one-shot animatable 3D full-head Gaussian avatars in a feed-forward manner.
    \item We propose a new framework for the novel setting of one-shot feed-forward 360$^\circ$ animatable avatar creation, which inpaints incomplete input local appearance details using global full-head priors from a pretrained 3D GAN in the UV space of a parametric face model for full-head Gaussian modeling.
    %, enabling 3D full-head modeling and avatar animation at the same time. 
    
    % generates Gaussians from the global full-head priors of a pretrained 3D GAN and the local appearance details of the input image in the UV space of a parametric face model, enabling 3D full-head modeling and animation at the same time.
    \item We perform symmetric inpainting on the local UV feature maps by taking advantage of the symmetric nature of human faces and the UV space, effectively enhancing the input appearance for high-fidelity 3D reconstruction.
    % \item We perform symmetric feature fusion on the local and global UV feature maps, effectively enhancing the global full-head prior with the local input details for high-fidelity 3D reconstruction.
    % \item We utilize a pretrained 3D GAN and its feed-forward inversion method for the full-head texture feature, and utilize the FLAME model for full-head geometry prior to deal with static occlusion.
    \item Extensive experiments on large-scale video datasets~\cite{xie2022vfhq, zhang2021flow} 
    \textcolor{black}{confirm} that our method \textcolor{black}{is the first to produce one-shot feed-forward 360$^\circ$ animatable avatars, with clear improvements in side and back view synthesis while achieving state-of-the-art performance in avatar animation.} %, improving the realism of 3D head avatars.
    %not only supports 360$^\circ$ view synthesis, but also outperforms state-of-the-art approaches in avatar animation, improving the realism of 3D head avatars.

\end{itemize}

\begin{figure*}[tb]
    \centering
  \includegraphics[width=1\textwidth]{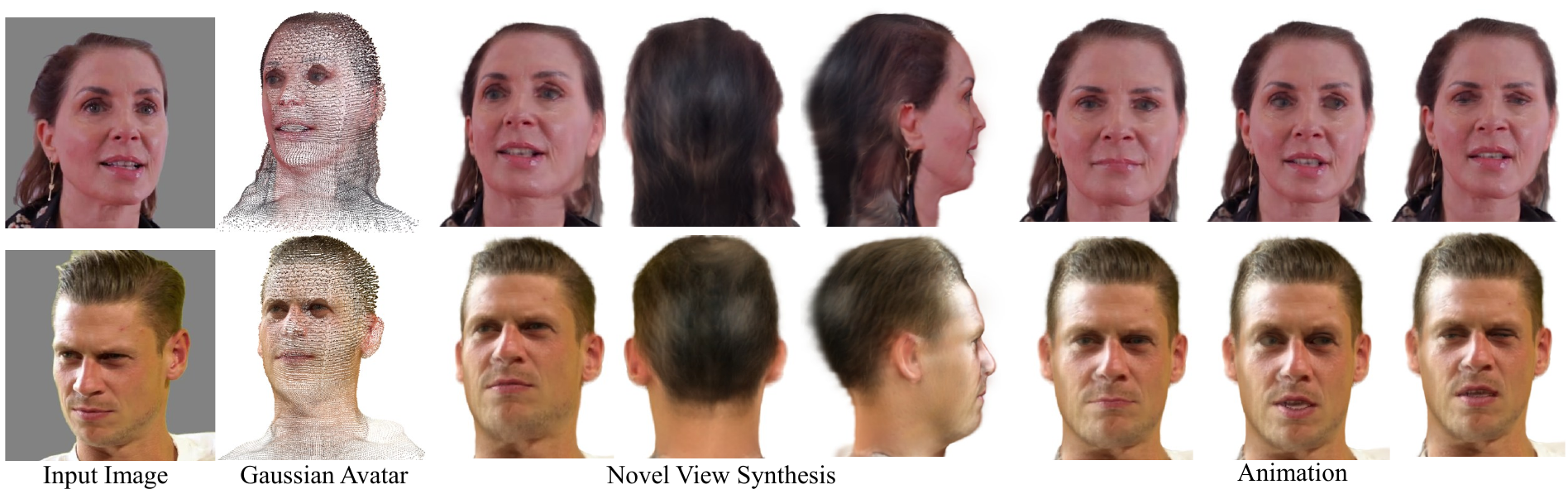}
  \vspace{-15pt}
  \caption{Given a single input image, our method reconstructs animatable 3D full-head Gaussian avatars in a single forward pass. These avatars provide consistent, high-fidelity 360° view synthesis and support real-time (246 FPS) animation.
  }
  \label{fig:teaser}
  \vspace{-6pt}
\end{figure*}
\section{Related Work}
\label{sec:related_work}
\vspace{-2pt}

\noindent\textbf{2D Talking Head Generation.} By leveraging powerful 2D generative models~\cite{goodfellow2014generative, ho2020denoising}, promising advancements have been made in 2D talking head generation by adding motion signals into them. Some works estimate motion in an unsupervised manner, representing it with warping fields described by unsupervised keypoints~\cite{siarohin2019first, hong2022depth, Zhao_2022_CVPR, tao2024learning, zhao2025synergizing} or latent features~\cite{wang2021latent, wang2023progressive}. Others adopt pretrained models to predict facial landmarks~\cite{zakharov2020fast, Ha_Kersner_Kim_Seo_Kim_2020, ma2024followyouremoji}. However, the lack of 3D face modeling hinders their performance, leading to face distortion and identity leakage. Some methods~\cite{Ren_2021_ICCV, Doukas_2021_ICCV, wang2021safa, yin2022styleheat} extract 3D Morphable Model (3DMM)~\cite{blanz1999morphable} parameters to introduce 3D information, but they cannot support arbitrary viewpoint rendering. In contrast, we directly model human heads in 3D to enable free-viewpoint rendering. %in $360^{\circ}$.

\noindent\textbf{Optimization-based 3D Animatable Head Avatar Reconstruction.} To achieve free-viewpoint rendering, a line of research reconstructs 3D animatable heads from multi-view or monocular videos of a specific person. By optimizing on videos of the same identity, these methods~\cite{qian2024gaussianavatars, zheng2023pointavatar, xiang2024flashavatar, zielonka2023instant, zhang2025fate, zheng2022avatar, li2025rgbavatar} generally achieve high visual quality. However, they require videos with thousands of frames for optimization, which hinders practical use. To reduce the need for lengthy video data, another line of research leverages strong priors from 3D GANs~\cite{chan2022efficient, an2023panohead, li2024spherehead, sun2023next3d} and diffusion models~\cite{rombach2022high}. Some methods~\cite{taubner2025cap4d, yin2025facecraft4d,zhou2025zero} generate multi-view data with different motions from a single image and optimize on the generated data. Others perform optimization-based GAN inversion~\cite{ma2023otavatar, roich2022pivotal, zielonka2025synthetic} from the input image on pretrained 3D GANs to create animatable 3D heads. However, the required optimization process remains time-consuming.
In contrast to these methods, we create one-shot 3D animatable heads in a feed-forward manner, avoiding the cumbersome optimization process.

\noindent\textbf{Feed-Forward 3D Animatable Head Avatar Reconstruction.}
To improve the generalizability, one-shot feed-forward 3D head reconstruction %from a single image in a single forward pass
has gained increasing attention. 
ROME~\cite{khakhulin2022realistic} extends 3DMM~\cite{li2017learning} meshes to model non-facial parts. With the excellent rendering quality of NeRF~\cite{mildenhall2020nerf}, some methods~\cite{li2023generalizable, tran2024voodoo, ye2024realdportrait, deng2024portrait4d, deng2024portrait4d_eccv, li2023one} incorporate motion features into efficient tri-planes~\cite{chan2022efficient} for high-fidelity head avatar creation. However, their rendering speed generally remains slow. In recent years, 3DGS~\cite{kerbl20233d} is widely leveraged for its extraordinary rendering quality and efficiency. GAGAvatar~\cite{chu2024generalizable} generates two sets of Gaussians for appearance and expression, and adds a neural renderer to enhance details. LAM~\cite{he2025lam} initializes Gaussians on FLAME~\cite{li2017learning} vertices and generates Gaussian attributes by querying image features through a transformer. However, trained on large monocular video datasets~\cite{xie2022vfhq, zhu2022celebv}, which rarely contain frames of the side and back of the head, these methods generally fail when the rendering viewpoint significantly differs from the input view. 
\textcolor{black}{Avat3r~\cite{kirschstein2025avat3r} and FastGHA~\cite{ji2026fastgha} leverage multi-view inputs and Vision Transformers to infer multi-view correspondence for high-quality reconstruction. On the contrary, we aim for 360$^{\circ}$ renderings from a single image.}
Recent FlexAvatar~\cite{kirschstein2025flexavatar} unifies training across monocular and multi-view datasets with learnable data source tokens, enabling the creation of complete 3D heads. 
In contrast, we directly inpaint the incomplete input local UV-space features with full-head priors from a pretrained 3D GAN~\cite{an2023panohead}, thus generating complete UV Gaussian attribute maps for full-head modeling.

\vspace{-3pt}
\section{Method}
\label{sec:method}

% \vspace{-10pt}
\vspace{-3pt}
\subsection{Overview}
\vspace{-3pt}
% \vspace{-5pt}

\begin{figure*}[tb]
    \centering
  \includegraphics[width=1\linewidth]{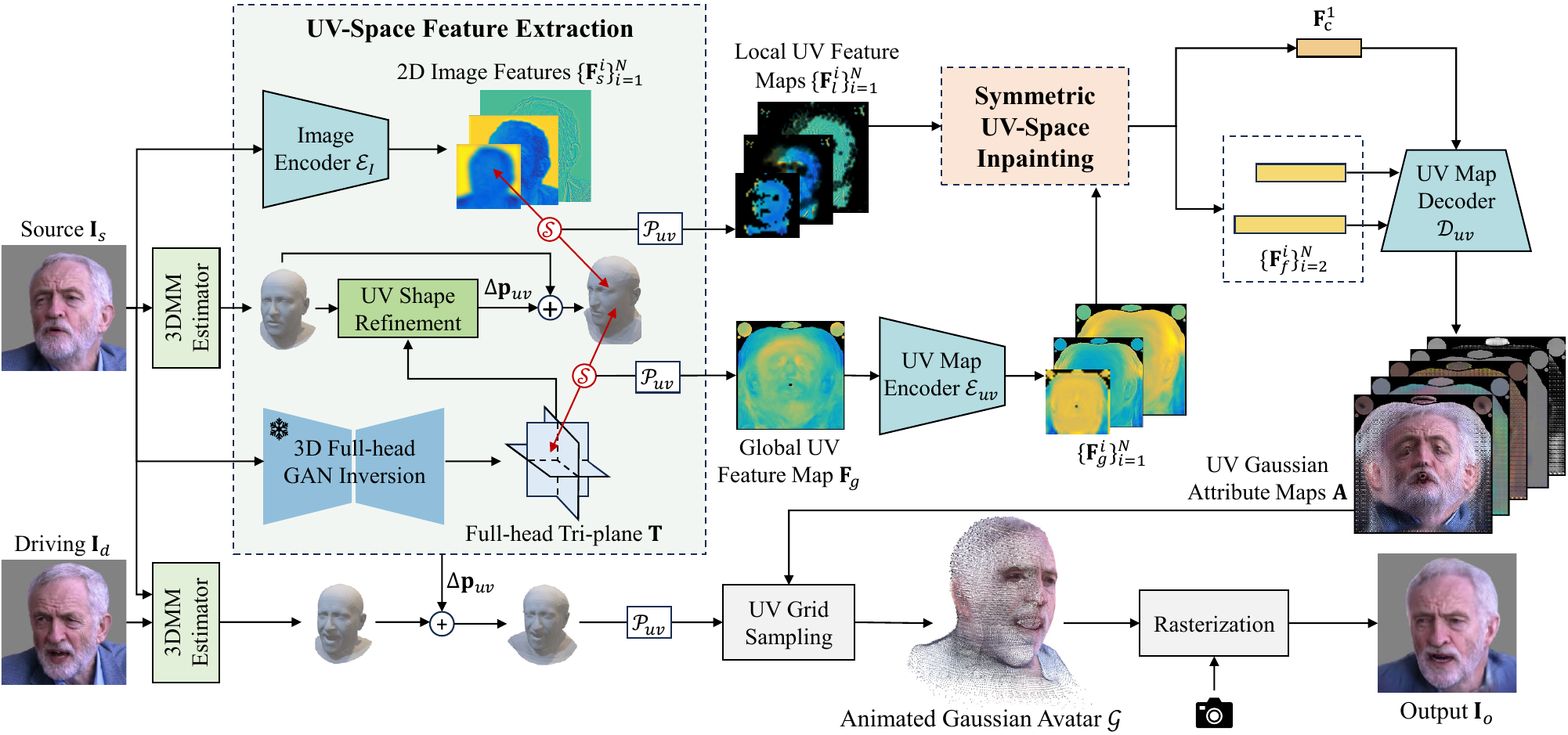}
  % \vspace{-10pt}
  \vspace{-16pt}
  \caption{Overview of the framework. Given an input source image, the UV-space feature extraction module extracts its global and local UV feature maps for animatable 3D full-head reconstruction. The symmetric UV-space inpainting module takes advantage of the symmetry of human faces and the UV space to inpaint the incomplete local UV feature maps with global features. From the predicted UV Gaussian attribute maps, 3D Gaussian primitives are sampled, which can be animated with a parametric face model and rendered given a camera pose.
  }
  \label{fig:overview}
  \vspace{-6pt}
\end{figure*}

Given an input source image of a human face, denoted as $\mathbf{I}_s$, \textcolor{black}{we aim} to reconstruct an animatable 3D full-head Gaussian avatar on the FLAME~\cite{li2017learning} mesh surface via UV parameterization. 
%Our framework is illustrated in Fig.~\ref{fig:overview}. 
\textcolor{black}{Fig.~\ref{fig:overview} depicts the overview of our framework.} 
First, the UV-space feature extraction module extracts the texture features from $\mathbf{I}_s$ and projects them into the UV space, producing a global UV feature map $\mathbf{F}_g$ and multi-scale local UV feature maps $\{\mathbf{F}_l^i\}_{i=1}^N$. It also predicts a shape offset $\Delta \mathbf{p}_{uv}$ to refine the estimated FLAME mesh shape based on $\mathbf{I}_s$. Next, the symmetric UV-space inpainting module takes advantage of the symmetric nature of human faces and the UV space, inpainting the incomplete but fine-grained local UV feature maps with global features to generate a set of UV Gaussian attribute maps, from which 3D Gaussian primitives are sampled to reconstruct the Gaussian avatar. Bound to the surface of the source FLAME mesh in canonical space, the Gaussian avatar can be animated with a novel FLAME expression derived from a driving image $\mathbf{I}_d$ and rendered given a camera pose to obtain the output image $\mathbf{I}_o$. Details on UV-space Gaussian modeling, the design of the two modules, the generation of UV Gaussian attribute maps, the regularization and training strategy are described in the following subsections.

%These 3D Gaussian primitives are bound to the surface of the source FLAME mesh in the canonical space, so the Gaussian avatar can be animated with a novel FLAME expression from a driving image $\mathbf{I}_d$ and rendered given a camera pose to obtain the output image $\mathbf{I}_r$.

%For animation, the FLAME mesh with a different expression is first deformed with $\Delta p_{uv}$. Then, 

\vspace{-3pt}
\subsection{UV-Space Gaussian Modeling}
\vspace{-3pt}

To utilize the animatable FLAME model for 3D full-head reconstruction and animation, we follow~\cite{kirschstein2024gghead, yu2025gaia} and bind 3D Gaussian primitives to the FLAME mesh surface via UV parameterization, enabling the use of efficient 2D backbones and regularizing Gaussian positions. Specifically, we generate a UV Gaussian attribute map $\mathbf{A}_{\star}\in \mathbb{R}^{K\times K\times d_{\star}}$ for each Gaussian attribute $\star \in \{\mathrm{color,rotation,} \allowbreak \mathrm{scale,opacity,position}\}$, where $K$ is the size of the UV map and $d_{\star}$ is the Gaussian attribute dimension, forming a set of UV Gaussian attribute maps $\mathbf{A} \in \mathbb{R}^{K\times K\times 14}$. Since each valid texel in the UV space already corresponds to a 3D position, we use $\mathbf{p} \in \mathbb{R}^{K\times K\times 3}$ to represent the FLAME-based UV-space 3D position map, and $\mathbf{A}_{\mathrm{position}}$ to represent the 3D position offset. 
%To make the scales of Gaussians change dynamically with the mesh faces that they are rigged to during animation, we compute the relative scales between the 3D face sizes in the driving mesh and their corresponding UV face sizes as a relative scaling map $\mathbf{s} \in \mathbb{R}^{K\times K\times 1}$. The final UV Gaussian scale map is updated as $\mathbf{A}_{\mathrm{scale}} = \mathbf{s} \odot \mathbf{A}_{\mathrm{scale}}$ where $\odot$ is the Hadamard product.
To balance the size differences between mesh faces in the UV space and in 3D space, we rescale the UV Gaussian scale map. Specifically, we compute the relative scales between the mesh face sizes on the 3D mesh surface and their corresponding UV face sizes as a relative scaling map $\mathbf{s} \in \mathbb{R}^{K\times K\times 1}$. The UV Gaussian scale map is further updated as $\mathbf{A}_{\mathrm{scale}} = \mathbf{s} \odot \mathbf{A}_{\mathrm{scale}}$, where $\odot$ is the Hadamard product.

To produce 3D Gaussian primitives $\mathcal{G}$, we perform grid sampling on $\mathbf{A}$ following~\cite{kirschstein2024gghead, yu2025gaia}. The process is formulated as:
\begin{equation}
\mathcal{G}= \mathrm{grid\_sample}(\mathbf{A},\mathcal{X}),
\end{equation}
where $\mathcal{X}$ is the set of sampling positions in the UV space.

\vspace{-3pt}
\subsection{UV-Space Feature Extraction}
\vspace{-2pt}

% \begin{figure}[tb]
%     \centering
%   \includegraphics[width=1\linewidth]{figures/uv_feat_extraction.pdf}
%   \caption{Illustration of UV Space Feature Extraction. We sample from the tri-plane of a pretrained 3D full-head GAN inversion method and 2D image features with points from a refined FLAME mesh surface to produce the global and local UV space feature maps.
%   }
%   \label{fig:uv_feat_extraction}
% \end{figure}

% To generate UV Gaussian attribute maps, we first extract the UV space features from the source image $\mathbf{I}_s$. As shown in Fig.~\ref{fig:uv_feat_extraction}, based on the sampled points in the UV space, we first refine the estimated FLAME mesh shape, then query their corresponding positions from a full-head tri-plane and 2D image features to produce global and local UV feature maps, which are integrated later.
To generate UV Gaussian attribute maps, we first extract the UV-space feature maps from the source image $\mathbf{I}_s$. The extracted UV feature maps are utilized for later integration.

\noindent\textbf{Full-Head Feature Generation.}
Since $\mathbf{I}_s$ only provides the head appearance from the source camera view, it is insufficient to reconstruct the full-head appearance using only information from $\mathbf{I}_s$. To obtain knowledge from the invisible areas in $\mathbf{I}_s$, we propose to utilize the full-head prior of a 3D full-head GAN named PanoHead~\cite{an2023panohead} and its feed-forward inversion method~\cite{bilecen2024dual} to map $\mathbf{I}_s$ to a full-head tri-plane $\mathbf{T}$, which provides coarse 3D full-head features for $\mathbf{I}_s$.

\noindent\textbf{UV Shape Refinement.}
Utilizing the source FLAME mesh predicted by the 3DMM estimator, we can directly sample the tri-plane feature for each 3D position corresponding to a UV-space texel, producing a UV tri-plane feature map $\mathbf{F_{T}^p}$. However, FLAME meshes often fail to model the shape of the hair, making the 3D position-based feature sampling inaccurate. We propose to use a 2D UNet $\mathcal{F}_{refine}$ to refine the mesh shape. Given $\mathbf{F_{T}^p}$ and the UV-space 3D position map $\mathbf{p}$, we predict the UV shape offset $\Delta \mathbf{p}_{uv}$ as:
\begin{equation}
\Delta \mathbf{p}_{uv}=\mathcal{F}_{refine}([\mathbf{F_{T}^p},\mathbf{p}]),
\end{equation}
where $[\cdot,\cdot]$ is the concatenation operation. Then, by adding $\Delta \mathbf{p}_{uv}$ to the original 3D position map $\mathbf{p}$, we obtain a refined 3D position map $\mathbf{p}_r=\mathbf{p}+\Delta \mathbf{p}_{uv}$.

%The FLAME meshes predicted by the 3DMM estimator can model the frontal face shape, but they often fail to model the shape of the hair. This makes it difficult to 

\noindent\textbf{Global UV Feature Extraction.}
With the refined 3D position map $\mathbf{p}_r$, we directly sample the full-head tri-plane $\mathbf{T}$ to produce the global UV feature map $\mathbf{F}_g$.
Produced by the pretrained 3D GAN with rich full-head prior knowledge, the tri-plane feature $\mathbf{T}$ contains convincing global head geometry and full-head textures for $\mathbf{I}_s$. Therefore, $\mathbf{F}_g$ contains a complete full-head representation of $\mathbf{I}_s$. However, constrained by the latent code dimension of the GAN inversion method and the resolution of the tri-plane, $\mathbf{T}$ cannot faithfully preserve the appearance details visible in $\mathbf{I}_s$, and consequently, neither can $\mathbf{F}_g$. %Still, $\mathbf{F}_g$ can serve as a full-head prior 
Additional local features from $\mathbf{I}_s$ are needed for high-fidelity avatar reconstruction.

% sample 6 more points and fuse them with an MLP?

\noindent\textbf{Local UV Feature Extraction.}
To extract appearance details from $\mathbf{I}_s$, we use a CNN-based encoder to encode $\mathbf{I}_s$ into multi-scale 2D image features $\{\mathbf{F}_s^i\}_{i=1}^N$. To map the image features to the UV space, we utilize the refined 3D position map $\mathbf{p}_r$. By projecting the 3D positions in $\mathbf{p}_r$ onto the 2D feature space according to the source camera pose, we can sample the corresponding 2D image feature for each UV texel, resulting in a set of multi-scale UV source features $\{\mathbf{F}_{s,uv}^i\}_{i=1}^N$. However, due to occlusion, only part of the 3D positions in $\mathbf{p}_r$ are visible in the source image. To avoid obtaining mismatched features in the UV space, we filter out invisible 3D positions in $\mathbf{p}_r$ following~\cite{hu2023sherf,xie2023high}, which results in a UV visibility mask $\mathbf{M}_v$. By applying this mask to $\{\mathbf{F}_{s,uv}^i\}_{i=1}^N$, we exclude textures sampled from the invisible positions in the source image. The resulting multi-scale local UV feature maps $\{\mathbf{F}_{l}^i\}_{i=1}^N$ preserve the local appearance details from $\mathbf{I}_s$, but they are incomplete for full-head modeling.

\vspace{-3pt}
\subsection{Symmetric UV-Space Inpainting}
\vspace{-2pt}

\begin{figure}[tb]
    \centering
  \includegraphics[width=0.9\linewidth]{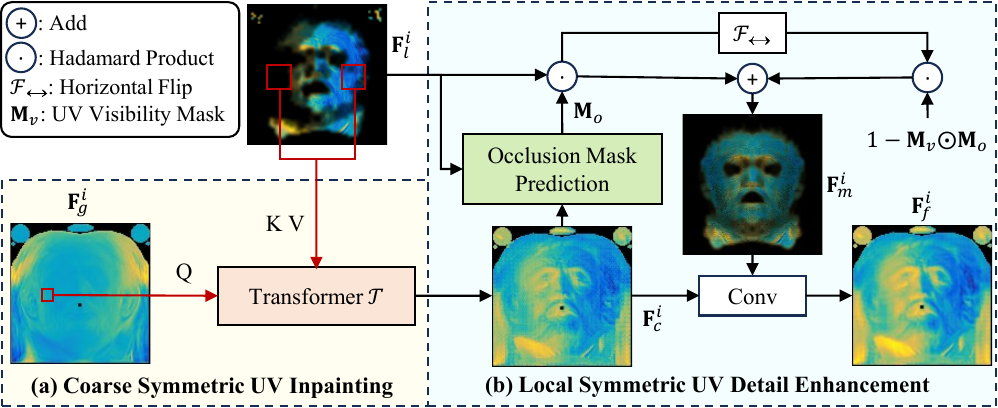}
  \caption{Illustration of symmetric UV-space inpainting at scale $i$. (a) Coarse Symmetric UV Inpainting: For each feature patch in $\mathbf{F}_{g}^i$, we query two symmetric local windows corresponding to the patch position in $\mathbf{F}_{l}^i$. (b) Local Symmetric UV Detail Enhancement: The output feature $\mathbf{F}_{c}^i$ from (a) is further enhanced by the local UV feature map and its symmetry with convolution.   %We query the  the encoded global UV feature map as query.
  }
  \vspace{-6pt}
  \label{fig:occ_comp}
\end{figure}

% We face two kinds of occlusion in 3D full-head animatable avatar modeling. In this module, we simultaneously fuse the global UV feature map and local UV feature maps while further compensating for occlusion using a UV codebook, dealing with both kinds of occlusion and producing the UV Gaussian attribute maps $\mathbf{A}$.

%Based on the extracted global and local UV feature maps, we 

% In this module, we aim to combine the coarse global UV feature map $\mathbf{F}_{g}$ with the fine-grained local UV feature maps $\{\mathbf{F}_{l}^i\}_{i=1}^N$ for UV Gaussian attribute map generation.
In this module, we aim to inpaint the fine-grained local UV feature maps $\{\mathbf{F}_{l}^i\}_{i=1}^N$ with the coarse but complete global UV feature map $\mathbf{F}_{g}$ for UV Gaussian attribute map generation.

To better inpaint the multi-scale local UV feature maps $\{\mathbf{F}_{l}^i\}_{i=1}^N$ with the global UV feature map $\mathbf{F}_{g}$, we also encode $\mathbf{F}_{g}$ with a CNN-based UV map encoder to produce multi-scale global UV feature maps $\{\mathbf{F}_{g}^i\}_{i=1}^N$, and inpaint $\mathbf{F}_{l}^i$ with $\mathbf{F}_{g}^i$  at scale $i$ across $N$ scales. Fig.~\ref{fig:occ_comp} illustrates symmetric UV-space inpainting at scale $i$.

%\noindent\textbf{Global and Local UV Feature Fusion at Scale i.}%Occlusion Compensation at scale i.}

% \noindent\textbf{Symmetric UV Space Inpainting at Scale i.}
\noindent\textbf{Coarse Symmetric UV Inpainting.}
% \subsubsection{Symmetric UV Space Feature Fusion at Scale i.}
% \noindent\textbf{Symmetric Window-based Cross Attention.}
%As a global full-head feature, $\mathbf{F}_{g}^i$ provides plausible texture for each valid texel in the UV space, even for the invisible area in the source view. However, it lacks the appearance details visible in the source image, and we need to transfer such details from $\mathbf{F}_{l}^i$ to it. Intuitively, $\mathbf{F}_{g}^i$ and $\mathbf{F}_{l}^i$ are inherently aligned, as each valid UV texel in $\mathbf{F}_{g}^i$ and $\mathbf{F}_{l}^i$ corresponds to the same 3D position on human heads. However, we generate the global and local UV feature maps using 3D position-based sampling and 3D projection, and the 3D positions and projections may still contain errors.
As a global full-head feature, $\mathbf{F}_{g}^i$ provides plausible texture for each valid texel in the UV space, even for the invisible area in the source view. However, it lacks the appearance details visible in the source image, which are contained in $\mathbf{F}_{l}^i$. Therefore, to inpaint $\mathbf{F}_{l}^i$, we can transfer appearance details from $\mathbf{F}_{l}^i$ to $\mathbf{F}_{g}^i$, which will result in a completed UV-space feature. Intuitively, $\mathbf{F}_{l}^i$ and $\mathbf{F}_{g}^i$ are inherently aligned, as each valid UV texel in $\mathbf{F}_{l}^i$ and $\mathbf{F}_{g}^i$ corresponds to the same 3D position on human heads. However, we generate the global and local UV feature maps using 3D position-based sampling and 3D projection, and the 3D positions and projections may still contain errors.
%Although we have one-to-one correspondence between $\mathbf{F}_{g}^i$ and $\mathbf{F}_{l}^i$ in the UV space, considering that mesh-based projection may still contain errors, 
To alleviate the potential misalignment issue caused by the errors, inspired by~\cite{pan2024humansplat}, we use a transformer architecture to fuse the features, where $\mathbf{F}_{g}^i$ serves as the query while $\mathbf{F}_{l}^i$ serves as the key and value at the cross-attention layer. As illustrated in Fig.~\ref{fig:occ_comp} (a), for each patch token in $\mathbf{F}_{g}^i$, we query tokens located in a local window of size $w\times w$ centered at the corresponding patch token in $\mathbf{F}_{l}^i$, allowing for small deviations in 3D positions and projections. However, if the source image contains a side view, tokens on the opposite side of the face in the UV space can hardly retrieve useful information from $\mathbf{F}_{l}^i$, as its corresponding local window only contains the masked-out area. Essentially, human faces are generally symmetric, and the face UV space is also symmetric. To better utilize the symmetry and retrieve more useful information from $\mathbf{F}_{l}^i$, we also use patch tokens in the symmetric local window of $\mathbf{F}_{l}^i$ as key-value pairs for cross-attention.

\noindent\textbf{Local Symmetric UV Detail Enhancement.}
As patchification inevitably leads to information loss, especially for large patches used in large-resolution feature maps at large scale $i$, appearance details contained in $\{\mathbf{F}_{l}^i\}_{i=2}^N$ may not be fully transferred to the transformer outputs $\{\mathbf{F}_{c}^i\}_{i=2}^N$. To further enhance the local appearance details from $\{\mathbf{F}_{l}^i\}_{i=2}^N$ within the aligned areas, we predict an occlusion mask $\mathbf{M}_o$ based on $\mathbf{F}_{c}^i$ and $\mathbf{F}_{l}^i$ to mask out the region of erroneous 3D projections in $\mathbf{F}_{l}^i$, producing a more accurate local feature $\mathbf{F}_{l,m}^i$:
\begin{equation}
\mathbf{F}_{l,m}^i=\mathbf{M}_o \odot \mathbf{F}_{l}^i, 
\end{equation}
where $\odot$ is the Hadamard product.
To utilize the symmetric nature of faces and the UV space, we horizontally flip the masked $\mathbf{F}_{l,m}^i$ and add it to the masked-out area in $\mathbf{F}_{l,m}^i$, taking full advantage of the appearance information from $\mathbf{F}_{l}^i$. The output $\mathbf{F}_{m}^i$ is further fused with $\mathbf{F}_{c}^i$ with a convolution layer to produce the final $\mathbf{F}_{f}^i$. The process is formulated as:
% \begin{equation}
% \begin{split}
% \mathbf{F}_{m}^i = \mathbf{F}_{l,m}^i+\mathrm{Flip}(\mathbf{F}_{l,m}^i) \cdot (1-\mathbf{M}_v \mathbf{M}_o),
% \end{split}
% \end{equation}
\begin{equation}
\mathbf{F}_{m}^i = \mathbf{F}_{l,m}^i+\mathcal{F}_{\leftrightarrow}(\mathbf{F}_{l,m}^i) \odot (1-\mathbf{M}_v \odot \mathbf{M}_o),
\end{equation}
\begin{equation}
\mathbf{F}_{f}^i = \mathrm{Conv}([\mathbf{F}_{c}^i,\mathbf{F}_{m}^i]),
\end{equation}
where $\mathcal{F}_{\leftrightarrow}$ is the horizontal flip operation, $\odot$ is the Hadamard product, and $[\cdot,\cdot]$ is the concatenation operation.

%\noindent\textbf{UV Gaussian Attribute Map Generation.}
\subsection{UV Gaussian Attribute Map Generation}
\vspace{-2pt}

We generate the set of UV Gaussian attribute maps $\mathbf{A}$ with the UV map decoder $\mathcal{D}_{uv}$ and the inpainted UV features $\mathbf{F}_{c}^1$ and $\{\mathbf{F}_{f}^i\}_{i=2}^N$. We use the transformer output $\mathbf{F}_{c}^1$ instead of $\mathbf{F}_{f}^1$, as $\mathbf{F}_{l}^1$ has the smallest resolution and contains few appearance details. $\mathbf{F}_{c}^1$ serves as the initial input to $\mathcal{D}_{uv}$, where it is gradually upsampled and processed with residual blocks. $\{\mathbf{F}_{f}^i\}_{i=2}^N$ is added to the intermediate feature maps with the same size in $\mathcal{D}_{uv}$. The final output of $\mathcal{D}_{uv}$ is $\mathbf{A}$. We also generate another attribute map $\mathbf{A}^1$ only using $\mathbf{F}_{c}^1$ as the input, which is also used to sample Gaussian attributes $\mathcal{G}^1$ and render an output image $\mathbf{I}_o^1$ during training, preventing $\mathcal{D}_{uv}$ from relying too much on high-resolution inputs.

\subsection{Regularization and Training Strategy}
\vspace{-2pt}

\begin{figure}[tb]
    \centering
  \includegraphics[width=0.8\linewidth]{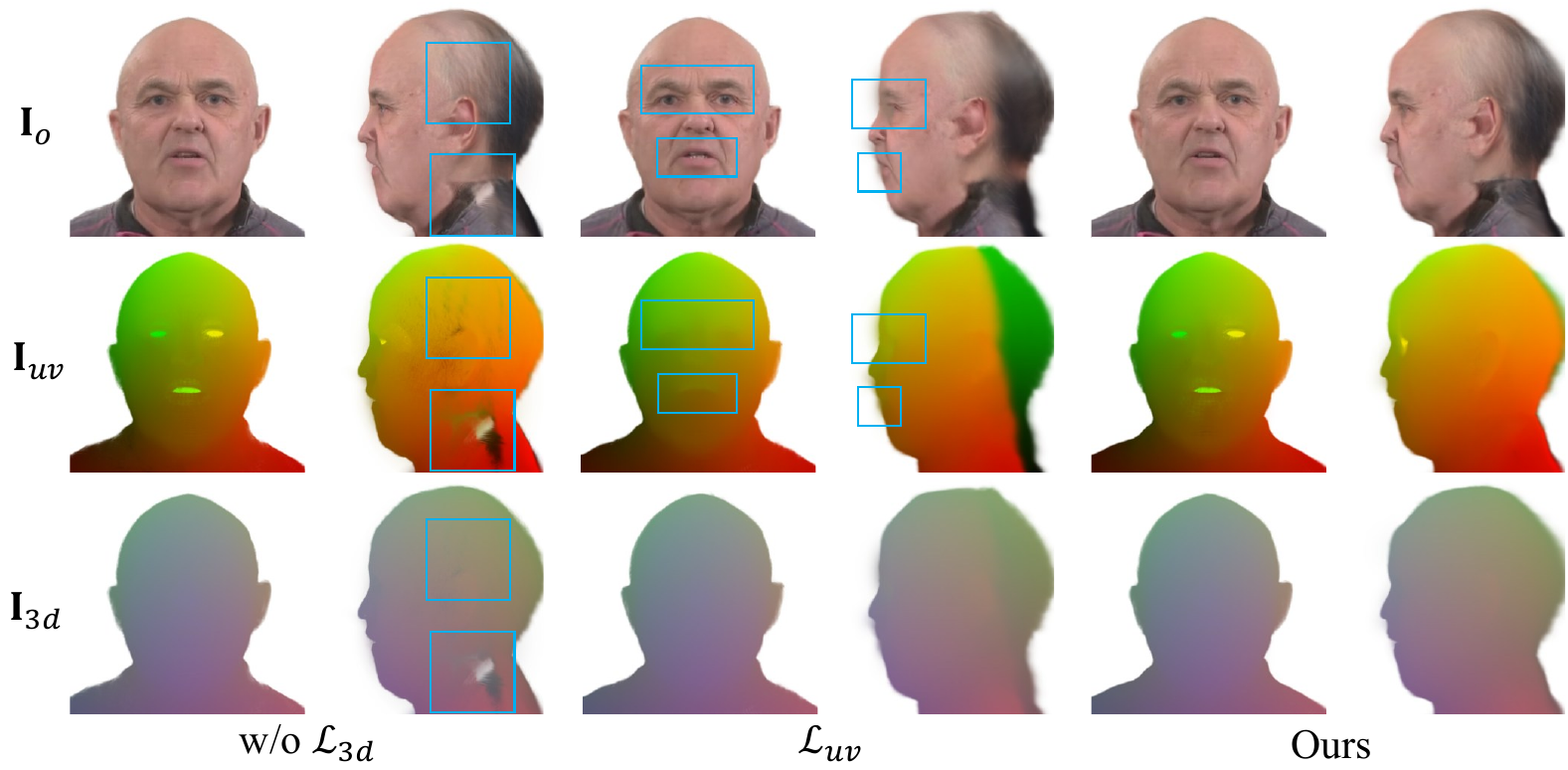}
  %\caption{Effect of the 3D total variation loss on the HDTF~\cite{wang2020mead} dataset. Blue boxes indicate erroneous areas in the rendered images.  }
  % \vspace{-6pt}
  \caption{Effect of the 3D total variation loss $\mathcal{L}_{3d}$. Compared with the UV total variation loss $\mathcal{L}_{uv}$ from~\cite{kirschstein2024gghead}, $\mathcal{L}_{3d}$ can alleviate holes on the avatar surfaces without bringing additional artifacts. Blue boxes indicate the erroneous areas in the rendered images.  }
  \label{fig:ablation_3dtv}
  \vspace{-6pt}
\end{figure}

\noindent\textbf{3D Total Variation Loss.}
As shown in Fig.~\ref{fig:ablation_3dtv}, we observe that Gaussian primitives modeled on the FLAME mesh surface normally cannot cover the whole avatar surface, resulting in holes that make the Gaussians representing the opposite side of the head visible, which harms 3D reconstruction quality. To alleviate such artifacts, inspired by the UV total variation (TV) loss in~\cite{kirschstein2024gghead} that uses TV loss on the UV renderings $\mathbf{I}_{uv}$ rendered from the same Gaussians $\mathcal{G}$ but with $\mathcal{G}_{\mathrm{color}}$ set to the UV coordinates, we propose to use the TV loss on the 3D renderings $\mathbf{I}_{3d}$, which is more suitable for the UV-space definitions where neighboring pixels in the rendered image can be far from each other in the UV space (\eg, eyeballs and eyelids). Specifically, based on the sampled Gaussians $\mathcal{G}$, we change the value of $\mathcal{G}_{\mathrm{color}}$ to $\mathcal{G}_{\mathrm{position}}$ while leaving other Gaussian attributes unchanged, and perform rendering to obtain the 3D rendering $\mathbf{I}_{3d}$. The 3D total variation loss is defined as:
\begin{equation}
\mathcal{L}_{3d}=\mathrm{TV}(\frac{\mathbf{I}_{3d}-(1-\mathbf{I}_{\alpha})}{\mathbf{I}_{\alpha}}),
\end{equation}
where $\mathbf{I}_\alpha$ is the rendered alpha map. 
This regularization encourages neighboring pixels in the rendered image to be rendered from neighboring Gaussians in the 3D space. 

% Thus, our regularization $\mathcal{L}_{reg}=\lambda_{eye} \mathcal{L}_{eye}+\lambda_{pos}\mathcal{L}_{pos}+\lambda_{shape}\mathcal{L}_{shape}+\lambda_{shape}^{tv} \mathcal{L}_{shape}^{tv}+\lambda_{3d} \mathcal{L}_{3d}$, where $\lambda_*$ is the regularization weight.

\noindent\textbf{Training Strategy.}
Except the offline 3DMM estimator and the frozen pretrained 3D full-head GAN and its inversion, we train the whole framework end-to-end. During training, we randomly \textcolor{black}{pick} two frames from a video of the same identity, in which one frame serves as the source image and the other serves as the driving image. With the source image $\mathbf{I}_s$ and the 3DMM motion parameter from the driving image $\mathbf{I}_d$, we aim to reconstruct a target image, which is $\mathbf{I}_d$. As large-scale face video datasets generally contain frontal faces, we rely on the pretrained 3D full-head GAN and its inversion to generate pseudo multi-view images of $\mathbf{I}_d$ for supervision. However, we find that the inversion method tends to generate inconsistent invisible parts of the head given different frames of a video, providing inconsistent multi-view supervision for different $\mathbf{I}_d$ given the same $\mathbf{I}_s$, which may hamper the 3D reconstruction quality. Therefore, we follow~\cite{deng2024portrait4d} to randomly switch between the animation mode and the 3D reconstruction mode. Specifically, in animation mode, \textcolor{black}{$\mathbf{I}_s$ serves as the source image and $\mathbf{I}_d$ as the target image}. In 3D reconstruction mode, we \textcolor{black}{assign} $\mathbf{I}_d$ as the source image, $\mathbf{I}_d$ together with its inverted novel views as the target images to avoid the inconsistency in the pseudo supervision.

For the training objective, we render RGB images $\mathbf{I}_o$ and $\mathbf{I}_o^1$, alpha maps $\mathbf{I}_{\alpha}$ and $\mathbf{I}_{\alpha}^1$ from $\mathcal{G}$ and $\mathcal{G}^1$, respectively. We supervise the RGB images using the target image(s) with a reconstruction loss $\mathcal{L}_{re}=\mathcal{L}_{1}+\mathcal{L}_{lpips}$, and supervise the alpha maps with $\mathcal{L}_{1}$ loss, denoted as $\mathcal{L}_\alpha$. When $\mathbf{I}_o$ contains frontal faces, we also use an ID similarity loss~\cite{deng2019arcface} $\mathcal{L}_{id}$. 
%use $\mathcal{L}_{1}$ loss and LPIPS loss~\cite{zhang2018unreasonable} on the RGB images, and use $\mathcal{L}_{1}$ loss on the alpha maps. 

Apart from the 3D total variation loss, we also apply a series of regularizations to constrain the Gaussian primitives for better rendering quality. Following~\cite{yu2025gaia}, we apply the eyeball TV loss $\mathcal{L}_{eye}$ and the position offset regularization $\mathcal{L}_{pos}=||\mathbf{A}_{\mathrm{position}}||_2$. To limit the extent of FLAME mesh offset, we regularize the predicted UV shape offset $\Delta \mathbf{p}_{uv}$:
\begin{equation}
\mathcal{L}_{shape}=\max(||\Delta \mathbf{p}_{uv}||_2-\epsilon
,0),
\end{equation}
where $\epsilon$ is a threshold.
To avoid producing Gaussian outliers, we further add a TV regularization to $\Delta \mathbf{p}_{uv}$:
\begin{equation}
\mathcal{L}_{shape}^{tv}=\mathrm{TV}(\Delta \mathbf{p}_{uv}).
\end{equation}
Thus, our regularization is defined as:
\begin{equation}
% \begin{split}
% \mathcal{L}_{reg}=&\lambda_{3d} \mathcal{L}_{3d} +\lambda_{eye} \mathcal{L}_{eye}+\lambda_{pos}\mathcal{L}_{pos}\\&+\lambda_{shape}\mathcal{L}_{shape}+\lambda_{shape}^{tv} \mathcal{L}_{shape}^{tv},
% \end{split}
\mathcal{L}_{reg}=\lambda_{3d} \mathcal{L}_{3d} +\lambda_{eye} \mathcal{L}_{eye}+\lambda_{pos}\mathcal{L}_{pos}+\lambda_{shape}\mathcal{L}_{shape}+\lambda_{shape}^{tv} \mathcal{L}_{shape}^{tv},
\end{equation}
where $\lambda_*$ is the regularization weight.

% \begin{equation}
% \hat{\mathcal{G}} \leftarrow \mathcal{G},
% \end{equation}
% \begin{equation}
% \hat{\mathcal{G}}_{\mathrm{color}} \leftarrow \hat{\mathcal{G}}_{\mathrm{position}},
% \end{equation}

To summarize, our training objective is:
\begin{equation}
\mathcal{L}=\mathcal{L}_{re}+\mathcal{L}_{\alpha}+\lambda_{id}\mathcal{L}_{id}+\lambda^1(\mathcal{L}_{lpips}^1+\mathcal{L}_{\alpha}^1)+\mathcal{L}_{reg},
\end{equation}
where $\lambda_{id}$ and $\lambda^1$ are the loss weights, and $\mathcal{L}_{lpips}^1$ and $\mathcal{L}_{\alpha}^1$ denote the losses for $\mathbf{I}_o^1$ and $\mathbf{I}_{\alpha}^1$, respectively.
\vspace{-7pt}
\section{Experiments}
\label{sec:exp}

\begin{figure*}[tb]
    \centering %\includegraphics[width=0.9\linewidth]{figures/overview_simple.pdf}
  \includegraphics[width=1\linewidth]{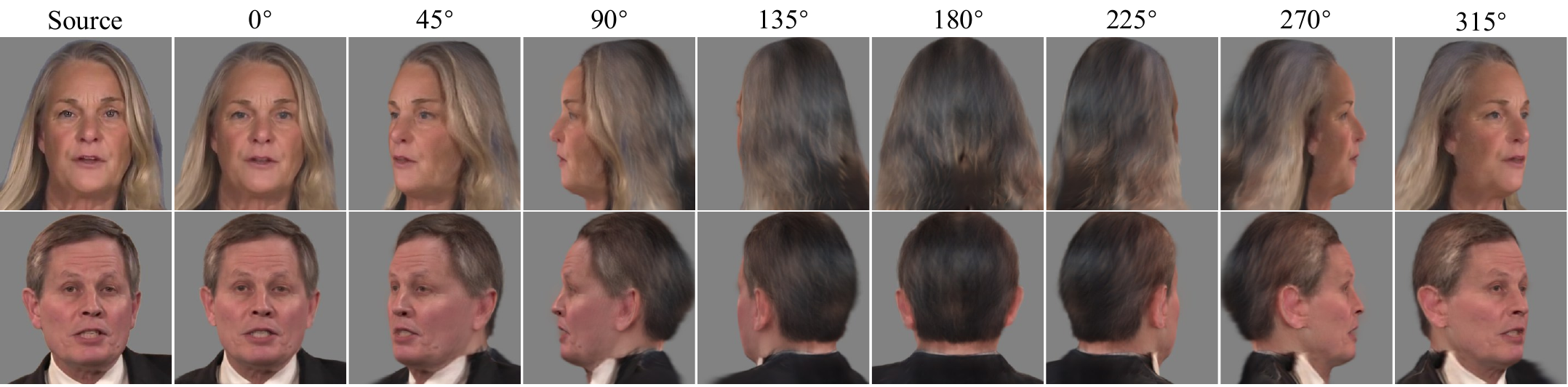}
  \vspace{-15pt}
  \caption{Multi-view results of our method on the HDTF~\cite{zhang2021flow} dataset. Our avatars can be viewed in 360$^\circ$.
  }
  \vspace{-5pt}
  \label{fig:multi_view}
\end{figure*}

\begin{table*}[!tb]
  %\begin{center}
  \caption{\textcolor{black}{Quantitative comparison of our method against state-of-the-art approaches} on the VFHQ dataset~\cite{xie2022vfhq}.}
  \vspace{-6pt}
  \centering
  \resizebox{1\textwidth}{!}{ %
  \begin{tabular}{l|ccccccc|ccc} %l}
  \toprule[1.2pt]
    \multirow{2}{*}{Method}  &  \multicolumn{7}{c|}{Self-reenactment}  &  \multicolumn{3}{c}{Cross-identity Reenactment} \\
    %\cline{2-7} \cline{8-13}
    
             & PSNR $\uparrow$  & SSIM $\uparrow$   & LPIPS $\downarrow$ & CSIM $\uparrow$  & AKD $\downarrow$ & AED $\downarrow$ & %APD $\downarrow$
             APD $\downarrow$
             & CSIM $\uparrow$  & AED $\downarrow$ & APD $\downarrow$\\
    \midrule
    StyleHEAT~\cite{yin2022styleheat} 
    & 18.96  & 0.7505  & 0.3821  & 0.1695  &  20.78  &  0.7236  &  %0.0681
    4.1380
    & 0.1096 & 0.8195 & 3.3018
    \\
    
    GOHA~\cite{li2023generalizable}
    & 19.87 & 0.7322  & 0.2769  &  0.6469  & 3.746   & 0.4279  & %0.0149
    1.0326
    & 0.4712 & 0.6859  &  1.9107
    
    \\

    Real3DPortrait~\cite{ye2024realdportrait}
    & 21.00 & 0.7572  & 0.2915  &  0.7696  &  3.934  & 0.3918  & %0.0185
    1.1890
    & 0.6562 & 0.7186  &  {1.7875 }
    \\

    % Portrait4D~\cite{deng2024portrait4d}
    % & 19.10 & 0.7343  &  0.3858 &    & 5.032  &   & 
    
    % &  &  & 
    % \\
    
    % Portrait4D-v2~\cite{deng2024portrait4d_eccv}
    % & 19.94 &  0.7590 & 0.3076  &    &  4.704  &   & 
    % & &   & 
    % \\

    Portrait4D~\cite{deng2024portrait4d}
    & 19.33 & 0.7166  &  0.3799 & 0.7339 & 6.239  & 0.4116  & 1.7931
    & 0.5812 & 0.7147 & 2.2928
    \\
    
    Portrait4D-v2~\cite{deng2024portrait4d_eccv}
    & 20.66 &  0.7421 & 0.2719  & 0.8075 & 5.367  & 0.3539 & 1.6181
    & 0.6728 & 0.6625 & 2.2002
    \\
    
    GAGAvatar~\cite{chu2024generalizable}
    & 21.60 & 0.7745  & \textbf{0.2249}  & \textbf{0.8459} &  {2.954}  & \textbf{0.3125}  & %\textbf{0.0119}
     {0.8898}
     & 0.6681 & \textbf{0.6409}  &  \textbf{1.7486}
    \\

    LAM~\cite{he2025lam}
    & {21.67} & {0.7756}  & 0.2716  &  0.6846  & 3.676   & 0.4535  & %0.0235
    1.4286
    & 0.5562 & 0.7271  & 3.2028
    \\
    
    Ours 
    & \textbf{23.24} & \textbf{0.7995}  & {0.2384} &  0.8012  &  \textbf{2.798}  & 0.3634  & %\underline{0.0139}
    \textbf{0.8660}
    & \textbf{0.6757} & 0.7103  & 2.3661
    \\ 

    \bottomrule[1.2pt]
  \end{tabular}
  }
  % \vspace{-7pt}
  \label{table:sota_vfhq}

  \vspace{-6pt}
\end{table*}

% \vspace{-3pt}
\subsection{Experimental Setup}
\vspace{-2pt}

\noindent\textbf{Datasets.}
We train our model on the VFHQ dataset~\cite{xie2022vfhq}. For data preprocessing, we follow~\cite{an2023panohead,bilecen2024dual} to estimate the camera pose, crop and remove the background of each video frame. All frames are resized to $512\times 512$. To obtain the FLAME parameters, we utilize the head tracker from~\cite{qian2024gaussianavatars}.
For performance evaluation, we use the VFHQ test split. To evaluate the generalizability, we randomly sample 20 videos from HDTF dataset~\cite{zhang2021flow} following~\cite{he2025lam, chu2024generalizable}. We also sample 21 videos from a multi-view MEAD~\cite{wang2020mead} video dataset to assess the effectiveness of our design under large camera pose changes.

\noindent\textbf{Metrics.}
For self-reenactment, given ground-truth frames, \textcolor{black}{we employ PSNR, SSIM, LPIPS~\cite{zhang2018unreasonable} to measure the reconstruction quality}. For identity preservation, we use cosine similarity (CSIM) of identity features extracted by~\cite{deng2019arcface}. For driving accuracy, we use average keypoint distance (AKD) based on~\cite{bulat2017far}, average expression distance (AED) % and average pose distance (APD) based on~\cite{retsinas20243d}.
based on~\cite{retsinas20243d} and average pose distance (APD) based on~\cite{lugaresi2019mediapipe}. For the setting of cross-identity reenactment, with no ground-truth available, we use CSIM, AED and APD for evaluation, following prior works~\cite{he2025lam, chu2024generalizable}.

\vspace{-5pt}
\subsection{Comparison with State-of-the-art Methods}
\vspace{-3pt}

% \begin{sloppypar}
We compare with 2D-based StyleHEAT~\cite{yin2022styleheat}, and 3D-based GOHA~\cite{li2023generalizable}, Real3DPortrait~\cite{ye2024realdportrait}, Portrait4D~\cite{deng2024portrait4d}, Portrait4D-v2~\cite{deng2024portrait4d_eccv}, GAGAvatar~\cite{chu2024generalizable} and LAM~\cite{he2025lam}.
% \end{sloppypar}

\noindent\textbf{Quantitative Comparison.} 
We \textcolor{black}{report} the results of the quantitative comparison with state-of-the-art methods on the VFHQ~\cite{xie2022vfhq} and HDTF~\cite{zhang2021flow} datasets in Tab.~\ref{table:sota_vfhq} and Tab.~\ref{table:sota_hdtf}. For self-reenactment, we \textcolor{black}{utilize the initial frame from each video} as the source image and \textcolor{black}{the remainder} as the driving images. \textcolor{black}{To perform} cross-identity reenactment, we randomly sample a frame from the video of a different identity as the source image. Our method generally achieves the best performance for self-reenactment and remains competitive for cross-identity reenactment. \textcolor{black}{With respect to} reconstruction quality (\ie, PSNR, SSIM), our method significantly outperforms previous methods on both datasets, highlighting the effectiveness of symmetric UV-space inpainting, where appearance details from the input source image are fully preserved for Gaussian attribute map generation. Our method also achieves the best AKD and APD for self-reenactment, which confirms the accuracy of expression and pose during animation. For cross-identity reenactment, our method outperforms others in identity preservation on the VFHQ dataset, where large differences in camera pose between source and driving image pairs occur most frequently. This indicates the superiority of modeling 3D full heads. In terms of motion accuracy for cross-identity reeanctment, we achieve slightly worse AED and APD. Based on the motion from the tracked FLAME parameters, the expression and pose accuracy of our method is limited by the tracking accuracy. Nevertheless, for AED and APD scores, our method outperforms LAM~\cite{he2025lam}, which shares the same FLAME head tracker with us. 

% \vspace{-2pt}

\begin{figure*}[tb]
    \centering
  \includegraphics[width=1\linewidth]{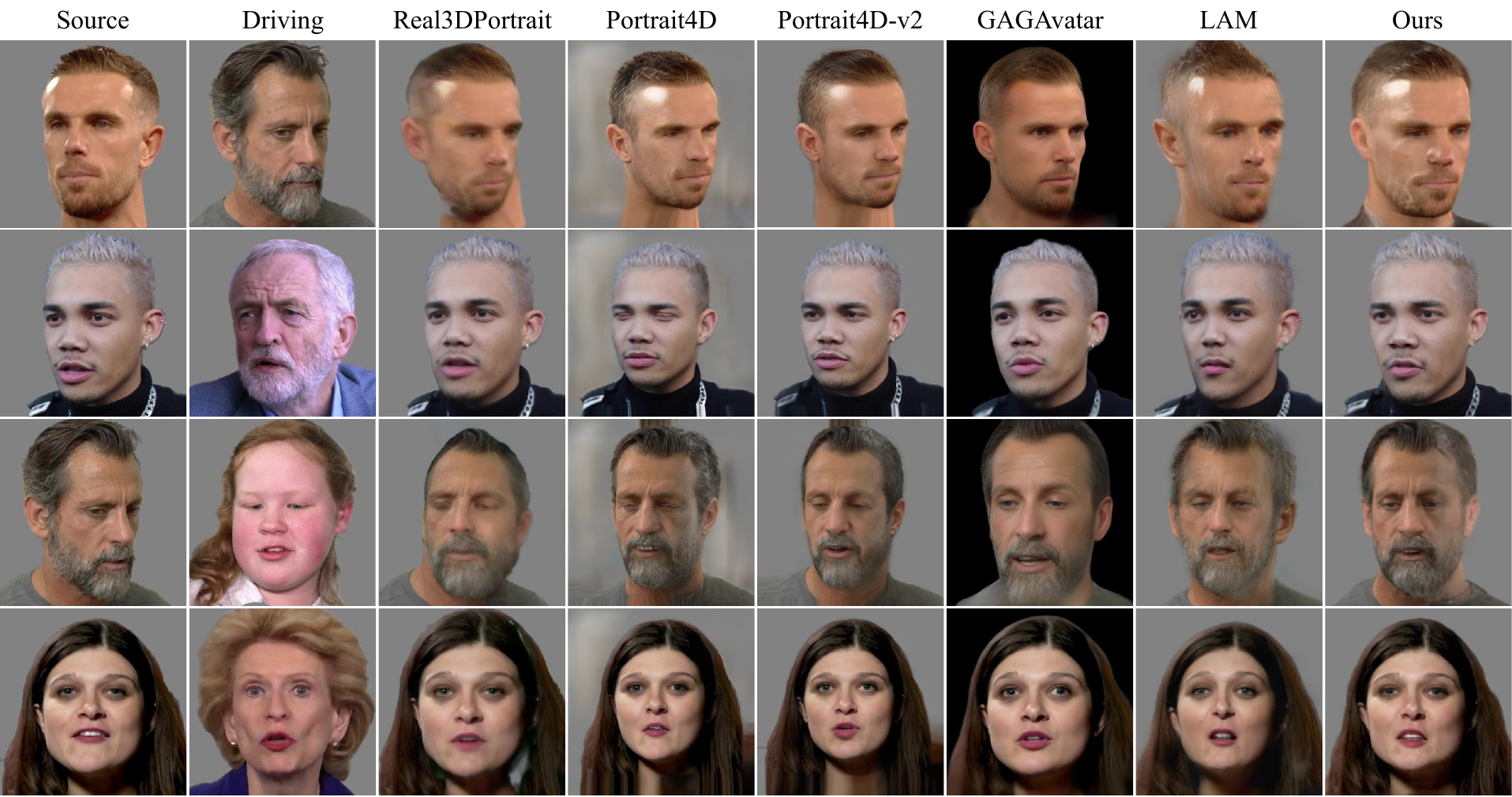}

  \vspace{-5pt}
  \caption{Qualitative comparison with state-of-the-art methods (\ie, Real3DPortrait~\cite{ye2024realdportrait}, Portrait4D~\cite{deng2024portrait4d}, Portrait4D-v2~\cite{deng2024portrait4d_eccv}, GAGAvatar~\cite{chu2024generalizable} and LAM~\cite{he2025lam}) for cross-identity reenactment on the VFHQ~\cite{xie2022vfhq} and HDTF~\cite{zhang2021flow} datasets. Our method best maintains source identity while effectively mimicking the driving motion.
  }
  \vspace{-2pt}
  \label{fig:sota_cmp_crossid}
\end{figure*}

%%%%%%%%%%%%%%%%%%%%%%%% HDTF
\begin{table*}[!tb]
  
  %\begin{center}
  \caption{\textcolor{black}{Quantitative comparison of our method against state-of-the-art approaches} on the HDTF dataset~\cite{zhang2021flow}.}
  \vspace{-6pt}
  \centering
  \resizebox{1\textwidth}{!}{ %
  \begin{tabular}{l|ccccccc|ccc} %l}
  \toprule[1.2pt]
    \multirow{2}{*}{Method}  &  \multicolumn{7}{c|}{Self-reenactment}  &  \multicolumn{3}{c}{Cross-identity Reenactment} \\
    %\cline{2-7} \cline{8-13}
    
             & PSNR $\uparrow$  & SSIM $\uparrow$   & LPIPS $\downarrow$ & CSIM $\uparrow$  & AKD $\downarrow$ & AED $\downarrow$ & %APD $\downarrow$
             APD $\downarrow$
             & CSIM $\uparrow$  & AED $\downarrow$ & APD $\downarrow$\\
    \midrule
    StyleHEAT~\cite{yin2022styleheat} 
    & 21.91 & 0.8102  & 0.2979  & 0.5096 & 6.564  & 0.4860  & 1.1539   
    & 0.4624  & 0.7656 & 1.4266
    \\
    
    GOHA~\cite{li2023generalizable}
    & 21.33 & 0.7739  & 0.2436  & 0.7541  & 3.164  & 0.4128  & 0.7296   
    & 0.7471  & \textbf{0.7287}  & 1.2370
    
    \\

    Real3DPortrait~\cite{ye2024realdportrait}
    & 23.39 & 0.8110  & 0.2362  & 0.8570  & 3.104  & 0.3413  & 0.7633   
    & {0.8886}  & 0.7487  & \textbf{1.1770}
    \\

    % Portrait4D~\cite{deng2024portrait4d}
    % & 20.03 & 0.7462  & 0.3602  & 0.8049  & 8.390  & 0.3787  & 1.1614  
    % & 0.7873  & 0.7482  & 1.4098
    % \\
    
    % Portrait4D-v2~\cite{deng2024portrait4d_eccv}
    % & 20.86 & 0.7581  & 0.2638  & {0.8640}  & 8.400  & 0.3398  & 1.1432   
    % & 0.8669  & 0.7332  & 1.3919
    % \\

    Portrait4D~\cite{deng2024portrait4d}
    & 21.36 & 0.7801 & 0.3245 & 0.8053 & 4.444 & 0.3785 & 1.1675 
    & 0.7876 & 0.7483 & 1.4278
    \\
    
    Portrait4D-v2~\cite{deng2024portrait4d_eccv}
    & 22.72 & 0.7985 & 0.2197 & 0.8641 & 4.278 & 0.3392 & 1.1729 
    & 0.8668 & 0.7333 & 1.4197
    \\
    
    GAGAvatar~\cite{chu2024generalizable}
    & {23.72} & {0.817}7  & {0.2089}  & \textbf{0.8894}  & {2.679 } & \textbf{0.3034}  & 0.5977   
    & \textbf{0.9004}  & {0.7300}  & {1.2045}
    \\

    LAM~\cite{he2025lam}
    & 23.55 & 0.8167  & 0.2362  & 0.7542  & 3.386  & 0.4180  & 0.8911   
    & 0.7656  & 0.7731  & 1.6635
    \\
    
    Ours 
    & \textbf{26.61} & \textbf{0.8642}  & \textbf{0.1900}  & 0.8622  & \textbf{2.287}  & {0.3318}   & \textbf{0.5286}   
    & 0.8568  & 0.7347  & 1.5605
    \\ 

    \bottomrule[1.2pt]
  \end{tabular}
  }
  % \vspace{-7pt}
  \label{table:sota_hdtf}
  \vspace{-6pt}
\end{table*}

\begin{table*}[tb]
    \caption{\textcolor{black}{Animation} speed measured in FPS. \textcolor{black}{We exclude the time for source avatar reconstruction and driving motion extraction that can be computed beforehand, and} take an average over 100 frames.
    %All results exclude the time for source avatar reconstruction and driving motion extraction that can be calculated beforehand. We take an average over 100 frames.
  }
  \vspace{-6pt}
  \centering
  \resizebox{0.9\linewidth}{!}{
  \begin{tabular}{l|cccccccc} 
  \toprule[1.2pt]
  Method  &  StyleHEAT~\cite{yin2022styleheat}   & 
  GOHA~\cite{li2023generalizable}
    & 

    Real3DPortrait~\cite{ye2024realdportrait}
     &
    
    Portrait4D-v2~\cite{deng2024portrait4d_eccv}
      & 
    
    GAGAvatar~\cite{chu2024generalizable}
      &

    LAM~\cite{he2025lam}
      &  
    Ours 
  
  \\
             
    \midrule
    FPS & 2.19 & 4.91 & 11.02 & 18.39 & 58.11 & 231.74 & \textbf{246.00}
    \\ 

    \bottomrule[1.2pt]
  \end{tabular}
  }
  \vspace{-8pt}
  % \vspace{-13pt}
  \label{table:speed_fps}
\end{table*}

\noindent\textbf{Qualitative Comparison.}
Fig.~\ref{fig:sota_cmp_crossid} \textcolor{black}{presents a qualitative comparison of our cross-identity reenactment results against state-of-the-art methods. Our method better maintains source face identity and head shape than previous methods under large camera pose changes (rows 1 and 3), while accurately imitating driving expressions (rows 2 and 4). This confirms the importance of 3D full-head reconstruction during animation.} The 360$^\circ$ rendering results of the avatars reconstructed by our method in Fig.~\ref{fig:multi_view} validate the effectiveness of our UV-space full-head modeling, allowing for more realistic 3D avatars. More qualitative comparison results are shown in the supplementary material.

\noindent\textbf{\textcolor{black}{Animation} Speed.}
Tab.~\ref{table:speed_fps} measures the time of reenactment during inference using an NVIDIA RTX 3090. Based on the efficient 3DGS representation, our method achieves an FPS of 246, outperforming previous methods. This confirms the efficiency of our method and highlights its potential real-time applications.

\vspace{-7pt}
\subsection{Ablation Study}
\label{subsec:exp-ablation}
\vspace{-5pt}

\begin{figure}[tb]
    \centering
   \includegraphics[width=0.9\linewidth]{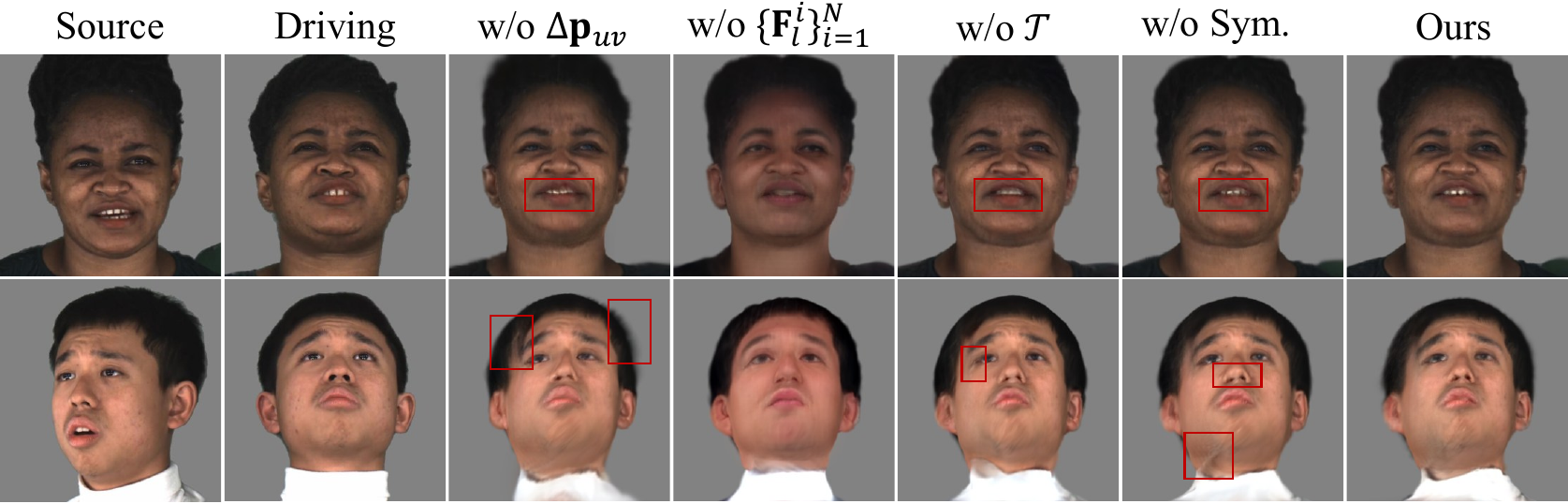}
   \vspace{-6pt}
  \caption{Qualitative ablation study on the MEAD~\cite{wang2020mead} dataset. Our method performs best with clearer facial features and less artifacts. %performs best under large pose changes.
  }

  \label{fig:ablation_design}
  \vspace{-2pt}
\end{figure}

\begin{figure}[tb]
    \centering
   \includegraphics[width=1\linewidth]{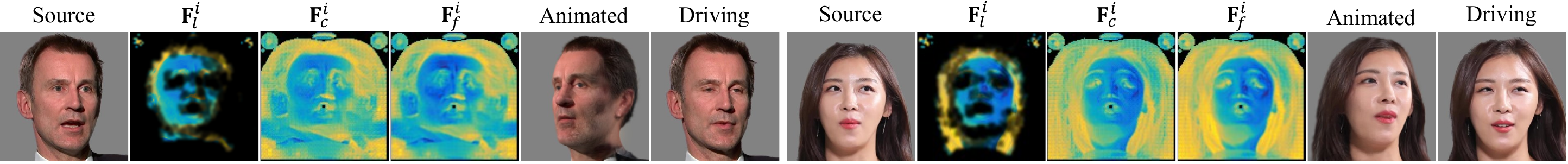}
   \vspace{-14pt}
  \caption{Visualization of the UV-space feature maps before and after symmetric UV-space inpainting. The inpainted UV-space features support 360$^\circ$ animatable avatars. We present results at scale $i=3$.
  }

  \label{fig:ablation_inpaint}
  \vspace{-6pt}
\end{figure}

We perform ablation studies to analyze the effectiveness of our framework design. The quantitative comparison is presented in Tab.~\ref{table:ablation_hdtf_} and the qualitative results are \textcolor{black}{depicted} in Fig.~\ref{fig:ablation_design}, Fig.~\ref{fig:ablation_inpaint} and Fig.~\ref{fig:ablation_3dtv}. 
 
\noindent\textbf{UV Shape Refinement.} 
According to Tab.~\ref{table:ablation_hdtf_}, removing UV shape refinement (w/o $\Delta \mathbf{p}_{uv}$) degrades the overall performance, as we cannot obtain a more accurate 3D position to sample the appropriate features from the full-head tri-plane and the 2D image features. Fig.~\ref{fig:ablation_design} shows blurry teeth and hair as well as clear boundary artifacts in the rendered frames without UV shape refinement.

\begin{table*}[tb]
    \caption{Ablation study on the design of our framework. We present the results for self-reenactment on the HDTF~\cite{zhang2021flow} and multi-view MEAD~\cite{wang2020mead} datasets. Our complete design achieves consistently the best results on both datasets.
  }
  \vspace{-6pt}
  \centering
  \resizebox{1\linewidth}{!}{
  \begin{tabular}{l|ccccccc|ccccccc} 
  \toprule[1.2pt]
  \multirow{2}{*}{Method}  &  \multicolumn{7}{c|}{HDTF}  &  \multicolumn{7}{c}{MEAD} \\
             & PSNR $\uparrow$  & SSIM $\uparrow$   & LPIPS $\downarrow$ & CSIM $\uparrow$  & AKD $\downarrow$ & AED $\downarrow$ & APD $\downarrow$
             & PSNR $\uparrow$  & SSIM $\uparrow$   & LPIPS $\downarrow$ & CSIM $\uparrow$  & AKD $\downarrow$ & AED $\downarrow$ & APD $\downarrow$
             \\
    \midrule
    w/o $\Delta \mathbf{p}_{uv}$
    & 26.15 & 0.8596 & 0.2058 & 0.8618 & 2.338 & 0.3321 & 0.5751
    & 17.09 & 0.7495 & 0.3493 & 0.6678 & 5.887 & 0.5032 & 2.6566
    \\

   w/o $\{\mathbf{F}_{l}^i\}_{i=1}^N$
    & 23.82 & 0.8186 & 0.2658 & 0.5230 & 2.732 & 0.3986 & 0.7115
    & 17.09 & \textbf{0.7508} & 0.3541 & 0.3932 & 4.849 & 0.6903 & 1.9517
    \\

    w/o $\mathcal{T}$
    & 25.79 & 0.8579 & 0.2049 & 0.8461 & 2.335 & 0.3443  & 0.5831
    & 17.05 & 0.7468 & 0.3430 & 0.6500 & \textbf{4.792} & 0.5044 & 2.1080
    \\

    w/o Sym. 
    & 26.60 & \textbf{0.8651} & \textbf{0.1893} & \textbf{0.8646} & 2.307 & 0.3402 & 0.5587
    & 17.18 & 0.7475 & \textbf{0.3402} & 0.6631 & 5.103 & 0.5047 & 2.2145
    \\
    % w/o $\mathcal{L}_{3d}$
    % & 26.60 & 0.8641 & \textbf{0.1890} & 0.8623 & \textbf{2.287} & \textbf{0.3238} & 0.5424
    % & 17.17 & 0.7474 & \textbf{0.3390} & \textbf{0.6878} & 4.680 & 0.5177 & 1.8844
    % \\
    $\mathcal{L}_{uv}$
    & 26.40 & 0.8622 & {0.1911} & 0.8567 & 2.324 & 0.3447 & 0.5370
    & 17.08 & 0.7459 & 0.3458 & 0.6663 & 5.058 & \textbf{0.4806} & 1.9627
    \\
    \midrule
    Ours 
    & \textbf{26.61} & {0.8642}  & {0.1900}  & 0.8622  & \textbf{2.287}  & \textbf{0.3318}   & \textbf{0.5286} 
    & \textbf{17.20} & {0.7495} & \textbf{0.3402} & \textbf{0.6765} & 4.844 & 0.5037 & \textbf{1.9484}
    \\ 

    \bottomrule[1.2pt]
  \end{tabular}
  }
  % \vspace{-8pt}
  
  \label{table:ablation_hdtf_}

  \vspace{-6pt}
\end{table*}

\noindent\textbf{Local UV Feature Maps.}
Tab.~\ref{table:ablation_hdtf_} and Fig.~\ref{fig:ablation_design} show that we can hardly preserve the identity from the input source image without using local UV feature maps (w/o $\{\mathbf{F}_{l}^i\}_{i=1}^N$), with a significant drop in CSIM. Although the full-head tri-plane is inverted from the source image, it cannot preserve appearance details, as the dimension of the inverted latent code and the tri-plane resolution are limited. We cannot reconstruct faithful 3D head avatars without using local UV feature maps, which confirms the importance of inpainting based on them.

\noindent\textbf{Symmetric UV-Space Inpainting.}
We utilize a transformer-based UV feature inpainting strategy with a symmetry prior. We analyze its effectiveness by removing the transformer (w/o $\mathcal{T}$) and removing symmetric operations (w/o Sym.) respectively. 

When the transformer is removed, the symmetric local UV features are directly fused with the global UV features without filtering out potential errors in 3D projection, which decreases the overall performance as shown in Tab.~\ref{table:ablation_hdtf_}. We can observe that the teeth are blurry and artifacts appear at the eye in Fig.~\ref{fig:ablation_design}, which is caused by inaccurate 3D projections. 

After we remove the symmetric operations, the quantitative results in Tab.~\ref{table:ablation_hdtf_} are generally comparable to the full design on HDTF, but drop consistently when evaluated on the multi-view MEAD dataset. Since the symmetric operations take the most effect when only a side face is visible in the source image, they cannot contribute significantly to the final results given the videos with frontal faces in HDTF. Nevertheless, they can generally improve full-head avatar animation quality on the multi-view video dataset where source and driving camera poses may vary significantly. As shown in Fig.~\ref{fig:ablation_design} row 2, the symmetric operations reduce artifacts at the source view boundaries, improving the reconstruction quality of the occluded parts. 
Additionally, Tab.~\ref{table:ablation_hdtf_} indicates that the symmetric operations consistently improve facial motion accuracy (\ie, AKD, AED, APD).
The reason is that by enforcing symmetry on local UV feature maps, the accuracy of projections from 2D image features to the UV space can be improved, making the generated Gaussian attribute maps more aligned with the underlying FLAME meshes. The improved projection accuracy is also confirmed by the clearer teeth when the symmetric operations are added in Fig.~\ref{fig:ablation_design} row 1.

Fig.~\ref{fig:ablation_inpaint} visualizes the UV-space feature maps before and after symmetric UV-space inpainting. Compared to the local UV map $\mathbf{F}_{l}^i$, the transformer output $\mathbf{F}_{c}^i$ provides a coarse but complete UV-space full-head representation. Further enhanced by $\mathbf{F}_{l}^i$, the final inpainting output $\mathbf{F}_{f}^i$ reduces grid-like artifacts caused by patchification in $\mathbf{F}_{c}^i$, preserving appearance details from the source view. The inpainted UV-space features are utilized for UV Gaussian attribute map generation, enabling 360$^\circ$ animatable Gaussian avatar reconstruction.
%the inpainted UV feature map $\mathbf{F}_{f}^i$ preserves a complete UV space full-head representation,  with appearance details from the source view, serving as the basis for 360$^\circ$ animatable avatar creation.

\noindent\textbf{3D Total Variation Loss.}
As shown in Fig.~\ref{fig:ablation_3dtv}, without using the 3D total variation loss (w/o $\mathcal{L}_{3d}$), the generated Gaussians cannot cover the avatar surface, leaving holes on the reconstructed avatars, which greatly harms the 3D quality. 

Using the UV total variation loss ($\mathcal{L}_{uv}$) from~\cite{kirschstein2024gghead} instead of $\mathcal{L}_{3d}$ can also encourage the Gaussians to cover the avatar surface, as $\mathcal{L}_{uv}$ encourages neighboring pixels in the rendered image to be from neighboring texels in the UV space. However, we use a different UV parametrization where eyeballs and inner mouth are far from the eyelid and lip areas in the UV space for full head modeling. Therefore, as shown in Fig.~\ref{fig:ablation_3dtv}, using $\mathcal{L}_{uv}$ leads to artifacts in the eye and mouth areas, causing Gaussians bound to other face areas to cover the eyeballs and mouth, which harms the avatar animation quality. The degradation is also confirmed by the quantitative results in Tab.~\ref{table:ablation_hdtf_}. In contrast, using our $\mathcal{L}_{3d}$ can realize a 3D consistent full-head avatar without damaging the animation quality.
\vspace{-5pt}
\section{Conclusion}
\vspace{-5pt}

% In this paper, we have tackled a novel setting of building one-shot 3D full-head animatable avatars in a feed-forward manner, which allows for more immersive 3D talking heads. We propose a novel framework that generates the Gaussian primitives in the UV space of a parametric head model for animation control. We leverage the rich 3D full-head prior knowledge from a pretrained 3D GAN, and perform symmetric UV-space inpainting on the incomplete but fine-grained input local image features with the extracted global full-head features, effectively improving 3D reconstruction fidelity. Extensive experiments confirm the effectiveness of our method, generating 3D head avatars that can be animated and viewed in 360$^\circ$.  

% In this paper, we have tackled a novel setting of building one-shot 3D full-head animatable avatars in a feed-forward manner, which allows for more immersive 3D talking heads. We propose a novel framework that generates complete Gaussian attribute maps in the UV space of a parametric head model via inpainting. Specifically, we perform symmetric UV-space inpainting on the incomplete but fine-grained input local image features with the rich 3D full-head prior knowledge from a pretrained 3D GAN, effectively improving the fidelity of 3D reconstruction. \textcolor{black}{Extensive experiments confirm the effectiveness of our method in generating one-shot feed-forward 3D animatable head avatars that can be viewed in 360$^\circ$}.

\textcolor{black}{In this paper, we have tackled a novel setting of building one-shot feed-forward 3D full-head animatable avatars}, which allows for more immersive 3D talking heads. We propose a novel framework that generates complete Gaussian attribute maps in the UV space of a parametric head model via inpainting. Specifically, we perform symmetric UV-space inpainting on the incomplete but fine-grained input local image features with the rich 3D full-head prior knowledge from a pretrained 3D GAN, \textcolor{black}{effectively enhancing the fidelity of 3D reconstruction. Extensive experiments confirm that our method is able to generate one-shot feed-forward 3D animatable head avatars that can be viewed in 360$^\circ$}.

% \begin{ack}
% Use unnumbered first level headings for the acknowledgments. All acknowledgments

% Do {\bf not} include this section in the anonymized submission, only in the final paper. You can use the \texttt{ack} environment provided in the style file to automatically hide this section in the anonymized submission.
% \end{ack}
\section*{Acknowledgements}
\vspace{-5pt}
The research is supported in part by Early Career Scheme of the Research Grants Council (RGC) of the Hong Kong SAR under grant No. 26202321, ITF PRP/046/24FX, Science \& Technology Cooperation Program of Shandong under grant No. SDST26EG01, SAIL Research Project, HKUST-Zeekr Coolaborative Research Fund.

% \section*{References}

% References follow the acknowledgments in the camera-ready paper. Use unnumbered first-level heading for
% the references. Any choice of citation style is acceptable as long as you are
% consistent. It is permissible to reduce the font size to \verb+small+ (9 point)
% when listing the references.
% Note that the Reference section does not count towards the page limit.
\medskip

{
\small

\bibliographystyle{splncs04}
\bibliography{main}

@String(CVPR= {IEEE Conf. Comput. Vis. Pattern Recog.})

@String(ICCV= {Int. Conf. Comput. Vis.})

@String(ECCV= {Eur. Conf. Comput. Vis.})

@String(NIPS= {Adv. Neural Inform. Process. Syst.})

@String(TOG= {ACM Trans. Graph.})

@String(ICLR = {Int. Conf. Learn. Represent.})

@String(AAAI = {AAAI})

@String(CVPRW= {IEEE Conf. Comput. Vis. Pattern Recog. Worksh.})

@String(CVPR  = {CVPR})

@String(ICCV  = {ICCV})

@String(ECCV  = {ECCV})

@String(NIPS  = {NeurIPS})

@String(TOG   = {ACM TOG})

@String(ICLR  = {ICLR})

@String(CVPRW= {CVPRW})

@inproceedings{qian2024gaussianavatars,
  title={Gaussianavatars: Photorealistic head avatars with rigged 3d gaussians},
  author={Qian, Shenhan and Kirschstein, Tobias and Schoneveld, Liam and Davoli, Davide and Giebenhain, Simon and Nie{\ss}ner, Matthias},
  booktitle={CVPR},
  year={2024}
}

@inproceedings{zheng2023pointavatar,
  title={Pointavatar: Deformable point-based head avatars from videos},
  author={Zheng, Yufeng and Yifan, Wang and Wetzstein, Gordon and Black, Michael J and Hilliges, Otmar},
  booktitle={CVPR},
  year={2023}
}

@inproceedings{xiang2024flashavatar,
  title={Flashavatar: High-fidelity head avatar with efficient gaussian embedding},
  author={Xiang, Jun and Gao, Xuan and Guo, Yudong and Zhang, Juyong},
  booktitle={CVPR},
  year={2024}
}

@inproceedings{zielonka2023instant,
  title={Instant volumetric head avatars},
  author={Zielonka, Wojciech and Bolkart, Timo and Thies, Justus},
  booktitle={CVPR},
  year={2023}
}

@inproceedings{li2023generalizable,
  title={Generalizable one-shot 3d neural head avatar},
  author={Li, Xueting and De Mello, Shalini and Liu, Sifei and Nagano, Koki and Iqbal, Umar and Kautz, Jan},
  booktitle=NIPS,
  year={2023}
}

@inproceedings{mildenhall2020nerf,
  title={NeRF: Representing Scenes as Neural Radiance Fields for View Synthesis},
  author={Mildenhall, Ben and Srinivasan, Pratul P and Tancik, Matthew and Barron, Jonathan T and Ramamoorthi, Ravi and Ng, Ren},
  booktitle={ECCV},
  year={2020},
}

@inproceedings{tran2024voodoo,
  title={Voodoo 3d: Volumetric portrait disentanglement for one-shot 3d head reenactment},
  author={Tran, Phong and Zakharov, Egor and Ho, Long-Nhat and Tran, Anh Tuan and Hu, Liwen and Li, Hao},
  booktitle={CVPR},
  year={2024}
}

@inproceedings{ye2024realdportrait,
title={Real3D-Portrait: One-shot Realistic 3D Talking Portrait Synthesis},
author={Zhenhui Ye and Tianyun Zhong and Yi Ren and Jiaqi Yang and Weichuang Li and Jiawei Huang and Ziyue Jiang and Jinzheng He and Rongjie Huang and Jinglin Liu and Chen Zhang and Xiang Yin and Zejun MA and Zhou Zhao},
booktitle={ICLR},
year={2024},
}

@inproceedings{deng2024portrait4d,
  title={Portrait4d: Learning one-shot 4d head avatar synthesis using synthetic data},
  author={Deng, Yu and Wang, Duomin and Ren, Xiaohang and Chen, Xingyu and Wang, Baoyuan},
  booktitle=CVPR,
  year={2024}
}

@inproceedings{deng2024portrait4d_eccv,
  title={Portrait4d-v2: Pseudo multi-view data creates better 4d head synthesizer},
  author={Deng, Yu and Wang, Duomin and Wang, Baoyuan},
  booktitle={ECCV},
  year={2024},
}

@article{kerbl20233d,
  title={3D Gaussian splatting for real-time radiance field rendering.},
  author={Kerbl, Bernhard and Kopanas, Georgios and Leimk{\"u}hler, Thomas and Drettakis, George},
  journal=TOG, 
  volume={42},
  number={4},
  pages={139--1},
  year={2023}
}

@inproceedings{he2025lam,
  title={LAM: Large Avatar Model for One-shot Animatable Gaussian Head},
  author={He, Yisheng and Gu, Xiaodong and Ye, Xiaodan and Xu, Chao and Zhao, Zhengyi and Dong, Yuan and Yuan, Weihao and Dong, Zilong and Bo, Liefeng},
  booktitle={SIGGRAPH},
  year={2025}
}

@inproceedings{chu2024generalizable,
  title={Generalizable and animatable gaussian head avatar},
  author={Chu, Xuangeng and Harada, Tatsuya},
  booktitle=NIPS,
  year={2024}
}

@article{guo2025sega,
  title={SEGA: Drivable 3D Gaussian Head Avatar from a Single Image},
  author={Guo, Chen and Su, Zhuo and Wang, Jian and Li, Shuang and Chang, Xu and Li, Zhaohu and Zhao, Yang and Wang, Guidong and Huang, Ruqi},
  journal={arXiv preprint arXiv:2504.14373},
  year={2025}
}

@article{li2017learning,
  title={Learning a model of facial shape and expression from 4D scans},
  author={Li, Tianye and Bolkart, Timo and Black, Michael J and Li, Hao and Romero, Javier},
  journal=TOG, 
  volume={36},
  number={6},
  pages={1--17},
  year={2017},
  publisher={Association for Computing Machinery (ACM)}
}

@inproceedings{goodfellow2014generative,
  title={Generative adversarial nets},
  author={Goodfellow, Ian J and Pouget-Abadie, Jean and Mirza, Mehdi and Xu, Bing and Warde-Farley, David and Ozair, Sherjil and Courville, Aaron and Bengio, Yoshua},
  booktitle=NIPS,
  year={2014}
}

@inproceedings{ho2020denoising,
  title={Denoising diffusion probabilistic models},
  author={Ho, Jonathan and Jain, Ajay and Abbeel, Pieter},
  booktitle=NIPS,
  year={2020}
}

@inproceedings{wang2021latent,
  title={Latent Image Animator: Learning to Animate Images via Latent Space Navigation},
  author={Wang, Yaohui and Yang, Di and Bremond, Francois and Dantcheva, Antitza},
  booktitle=ICLR,
  year={2022}
}

@inproceedings{wang2023progressive,
  title={Progressive disentangled representation learning for fine-grained controllable talking head synthesis},
  author={Wang, Duomin and Deng, Yu and Yin, Zixin and Shum, Heung-Yeung and Wang, Baoyuan},
  booktitle=CVPR,
  year={2023}
}

@inproceedings{siarohin2019first,
  title={First order motion model for image animation},
  author={Siarohin, Aliaksandr and Lathuili{\`e}re, St{\'e}phane and Tulyakov, Sergey and Ricci, Elisa and Sebe, Nicu},
  booktitle=NIPS,
  year={2019}
}

@inproceedings{hong2022depth,
  title={Depth-aware generative adversarial network for talking head video generation},
  author={Hong, Fa-Ting and Zhang, Longhao and Shen, Li and Xu, Dan},
  booktitle=CVPR,
  year={2022}
}

@inproceedings{zhao2025synergizing,
  title={Synergizing motion and appearance: Multi-scale compensatory codebooks for talking head video generation},
  author={Zhao, Shuling and Hong, Fa-Ting and Huang, Xiaoshui and Xu, Dan},
  booktitle=CVPR,
  year={2025}
}

@inproceedings{ma2024followyouremoji,
title={Follow-Your-Emoji: Fine-Controllable and Expressive Freestyle Portrait Animation},
  author={Ma, Yue and Liu, Hongyu and Wang, Hongfa and Pan, Heng and He, Yingqing and Yuan, Junkun and Zeng, Ailing and Cai, Chengfei and Shum, Heung-Yeung and Liu, Wei and others},
  booktitle={SIGGRAPH Asia},
  year={2024}
}

@inproceedings{zakharov2020fast,
  title={Fast bi-layer neural synthesis of one-shot realistic head avatars},
  author={Zakharov, Egor and Ivakhnenko, Aleksei and Shysheya, Aliaksandra and Lempitsky, Victor},
  booktitle=ECCV,
  year={2020},
}

@inproceedings{Ha_Kersner_Kim_Seo_Kim_2020, 
title={MarioNETte: Few-Shot Face Reenactment Preserving Identity of Unseen Targets}, 
booktitle=AAAI, 
author={Ha, Sungjoo and Kersner, Martin and Kim, Beomsu and Seo, Seokjun and Kim, Dongyoung}, 
year={2020}
}

@InProceedings{Zhao_2022_CVPR,
    author    = {Zhao, Jian and Zhang, Hui},
    title     = {Thin-Plate Spline Motion Model for Image Animation},
    booktitle = CVPR,
    year      = {2022},
}

@inproceedings{tao2024learning,
  title={Learning Motion Refinement for Unsupervised Face Animation},
  author={Tao, Jiale and Gu, Shuhang and Li, Wen and Duan, Lixin},
  booktitle=NIPS,
  year={2024}
}

@inproceedings{blanz1999morphable,
  title={A Morphable Model for the Synthesis of 3D Faces},
  author={Blanz, Volker and Vetter, Thomas},
  booktitle={SIGGRAPH},
  year={1999},
}

@InProceedings{Ren_2021_ICCV,
    author    = {Ren, Yurui and Li, Ge and Chen, Yuanqi and Li, Thomas H. and Liu, Shan},
    title     = {PIRenderer: Controllable Portrait Image Generation via Semantic Neural Rendering},
    booktitle = ICCV,
    year      = {2021},
}

@InProceedings{Doukas_2021_ICCV,
    author    = {Doukas, Michail Christos and Zafeiriou, Stefanos and Sharmanska, Viktoriia},
    title     = {HeadGAN: One-Shot Neural Head Synthesis and Editing},
    booktitle = ICCV,
    year      = {2021},
}

@inproceedings{wang2021safa,
  title={Safa: Structure aware face animation},
  author={Wang, Qiulin and Zhang, Lu and Li, Bo},
  booktitle={3DV},
  year={2021},
}

@inproceedings{zhang2025fate,
  title={Fate: Full-head gaussian avatar with textural editing from monocular video},
  author={Zhang, Jiawei and Wu, Zijian and Liang, Zhiyang and Gong, Yicheng and Hu, Dongfang and Yao, Yao and Cao, Xun and Zhu, Hao},
  booktitle=CVPR,
  year={2025}
}

@inproceedings{zheng2022avatar,
  title={Im avatar: Implicit morphable head avatars from videos},
  author={Zheng, Yufeng and Abrevaya, Victoria Fern{\'a}ndez and B{\"u}hler, Marcel C and Chen, Xu and Black, Michael J and Hilliges, Otmar},
  booktitle=CVPR,
  year={2022}
}

@inproceedings{khakhulin2022realistic,
  title={Realistic one-shot mesh-based head avatars},
  author={Khakhulin, Taras and Sklyarova, Vanessa and Lempitsky, Victor and Zakharov, Egor},
  booktitle=ECCV,
  year={2022},
}

@inproceedings{ma2023otavatar,
  title={Otavatar: One-shot talking face avatar with controllable tri-plane rendering},
  author={Ma, Zhiyuan and Zhu, Xiangyu and Qi, Guo-Jun and Lei, Zhen and Zhang, Lei},
  booktitle=CVPR,
  year={2023}
}

@inproceedings{chan2022efficient,
  title={Efficient geometry-aware 3d generative adversarial networks},
  author={Chan, Eric R and Lin, Connor Z and Chan, Matthew A and Nagano, Koki and Pan, Boxiao and De Mello, Shalini and Gallo, Orazio and Guibas, Leonidas J and Tremblay, Jonathan and Khamis, Sameh and others},
  booktitle=CVPR,
  year={2022}
}

@inproceedings{sun2023next3d,
  title={Next3d: Generative neural texture rasterization for 3d-aware head avatars},
  author={Sun, Jingxiang and Wang, Xuan and Wang, Lizhen and Li, Xiaoyu and Zhang, Yong and Zhang, Hongwen and Liu, Yebin},
  booktitle=CVPR,
  year={2023}
}

@article{roich2022pivotal,
  title={Pivotal tuning for latent-based editing of real images},
  author={Roich, Daniel and Mokady, Ron and Bermano, Amit H and Cohen-Or, Daniel},
  journal=TOG,
  volume={42},
  number={1},
  pages={1--13},
  year={2022},
  publisher={ACM New York, NY}
}

@inproceedings{li2025rgbavatar,
  title={RGBAvatar: Reduced Gaussian Blendshapes for Online Modeling of Head Avatars},
  author={Li, Linzhou and Li, Yumeng and Weng, Yanlin and Zheng, Youyi and Zhou, Kun},
  booktitle=CVPR,
  year={2025}
}

@inproceedings{li2023one,
  title={One-shot high-fidelity talking-head synthesis with deformable neural radiance field},
  author={Li, Weichuang and Zhang, Longhao and Wang, Dong and Zhao, Bin and Wang, Zhigang and Chen, Mulin and Zhang, Bang and Wang, Zhongjian and Bo, Liefeng and Li, Xuelong},
  booktitle=CVPR,
  year={2023}
}

@inproceedings{xie2022vfhq,
  title={Vfhq: A high-quality dataset and benchmark for video face super-resolution},
  author={Xie, Liangbin and Wang, Xintao and Zhang, Honglun and Dong, Chao and Shan, Ying},
  booktitle=CVPRW,
  year={2022}
}

@inproceedings{an2023panohead,
  title={Panohead: Geometry-aware 3d full-head synthesis in 360deg},
  author={An, Sizhe and Xu, Hongyi and Shi, Yichun and Song, Guoxian and Ogras, Umit Y and Luo, Linjie},
  booktitle=CVPR,
  year={2023}
}

@inproceedings{bilecen2024dual,
  title={Dual encoder GAN inversion for high-fidelity 3D head reconstruction from single images},
  author={Bilecen, Bahri Batuhan and G{\"o}kmen, Ahmet and Dundar, Aysegul},
  booktitle=NIPS,
  year={2024}
}

@inproceedings{li2024spherehead,
  title={Spherehead: stable 3d full-head synthesis with spherical tri-plane representation},
  author={Li, Heyuan and Chen, Ce and Shi, Tianhao and Qiu, Yuda and An, Sizhe and Chen, Guanying and Han, Xiaoguang},
  booktitle=ECCV,
  year={2024},
}

@inproceedings{kirschstein2024gghead,
  title={Gghead: Fast and generalizable 3d gaussian heads},
  author={Kirschstein, Tobias and Giebenhain, Simon and Tang, Jiapeng and Georgopoulos, Markos and Nie{\ss}ner, Matthias},
  booktitle={SIGGRAPH Asia},
  year={2024}
}

@inproceedings{xie2023high,
  title={High-fidelity 3d gan inversion by pseudo-multi-view optimization},
  author={Xie, Jiaxin and Ouyang, Hao and Piao, Jingtan and Lei, Chenyang and Chen, Qifeng},
  booktitle=CVPR,
  year={2023}
}

@inproceedings{yu2025gaia,
  title={GAIA: Generative Animatable Interactive Avatars with Expression-conditioned Gaussians},
  author={Yu, Zhengming and Li, Tianye and Sun, Jingxiang and Shapira, Omer and Park, Seonwook and Stengel, Michael and Chan, Matthew and Li, Xin and Wang, Wenping and Nagano, Koki and others},
  booktitle={SIGGRAPH},
  year={2025}
}

@inproceedings{hu2023sherf,
  title={Sherf: Generalizable human nerf from a single image},
  author={Hu, Shoukang and Hong, Fangzhou and Pan, Liang and Mei, Haiyi and Yang, Lei and Liu, Ziwei},
  booktitle=ICCV,
  year={2023}
}

@inproceedings{zhang2018unreasonable,
  title={The unreasonable effectiveness of deep features as a perceptual metric},
  author={Zhang, Richard and Isola, Phillip and Efros, Alexei A and Shechtman, Eli and Wang, Oliver},
  booktitle=CVPR,
  year={2018}
}

@inproceedings{deng2019arcface,
  title={Arcface: Additive angular margin loss for deep face recognition},
  author={Deng, Jiankang and Guo, Jia and Xue, Niannan and Zafeiriou, Stefanos},
  booktitle=CVPR,
  year={2019}
}

@inproceedings{wang2020mead,
  title={Mead: A large-scale audio-visual dataset for emotional talking-face generation},
  author={Wang, Kaisiyuan and Wu, Qianyi and Song, Linsen and Yang, Zhuoqian and Wu, Wayne and Qian, Chen and He, Ran and Qiao, Yu and Loy, Chen Change},
  booktitle=ECCV,
  year={2020},
}

@inproceedings{yin2022styleheat,
  title={Styleheat: One-shot high-resolution editable talking face generation via pre-trained stylegan},
  author={Yin, Fei and Zhang, Yong and Cun, Xiaodong and Cao, Mingdeng and Fan, Yanbo and Wang, Xuan and Bai, Qingyan and Wu, Baoyuan and Wang, Jue and Yang, Yujiu},
  booktitle=ECCV,
  year={2022},
}

@inproceedings{bulat2017far,
  title={How far are we from solving the 2d \& 3d face alignment problem?(and a dataset of 230,000 3d facial landmarks)},
  author={Bulat, Adrian and Tzimiropoulos, Georgios},
  booktitle=ICCV,
  year={2017}
}

@inproceedings{retsinas20243d,
  title={3d facial expressions through analysis-by-neural-synthesis},
  author={Retsinas, George and Filntisis, Panagiotis P and Danecek, Radek and Abrevaya, Victoria F and Roussos, Anastasios and Bolkart, Timo and Maragos, Petros},
  booktitle=CVPR,
  year={2024}
}

@article{lugaresi2019mediapipe,
  title={Mediapipe: A framework for building perception pipelines},
  author={Lugaresi, Camillo and Tang, Jiuqiang and Nash, Hadon and McClanahan, Chris and Uboweja, Esha and Hays, Michael and Zhang, Fan and Chang, Chuo-Ling and Yong, Ming Guang and Lee, Juhyun and others},
  journal={arXiv preprint arXiv:1906.08172},
  year={2019}
}

@inproceedings{zhang2021flow,
  title={Flow-guided one-shot talking face generation with a high-resolution audio-visual dataset},
  author={Zhang, Zhimeng and Li, Lincheng and Ding, Yu and Fan, Changjie},
  booktitle=CVPR,
  year={2021}
}

@inproceedings{pan2024humansplat,
  title={Humansplat: Generalizable single-image human gaussian splatting with structure priors},
  author={Pan, Panwang and Su, Zhuo and Lin, Chenguo and Fan, Zhen and Zhang, Yongjie and Li, Zeming and Shen, Tingting and Mu, Yadong and Liu, Yebin},
  booktitle=NIPS,
  year={2024}
}

@inproceedings{zhu2022celebv,
  title={CelebV-HQ: A large-scale video facial attributes dataset},
  author={Zhu, Hao and Wu, Wayne and Zhu, Wentao and Jiang, Liming and Tang, Siwei and Zhang, Li and Liu, Ziwei and Loy, Chen Change},
  booktitle=ECCV,
  year={2022},
}

@inproceedings{taubner2025cap4d,
  title={Cap4d: Creating animatable 4d portrait avatars with morphable multi-view diffusion models},
  author={Taubner, Felix and Zhang, Ruihang and Tuli, Mathieu and Lindell, David B},
  booktitle=CVPR,
  year={2025},
}

@inproceedings{rombach2022high,
  title={High-resolution image synthesis with latent diffusion models},
  author={Rombach, Robin and Blattmann, Andreas and Lorenz, Dominik and Esser, Patrick and Ommer, Bj{\"o}rn},
  booktitle=CVPR,
  year={2022}
}

@inproceedings{yin2025facecraft4d,
  title={FaceCraft4D: Animated 3D Facial Avatar Generation from a Single Image},
  author={Yin, Fei and Yao, Chun-Han and Mantiuk, Rafal K and Jampani, Varun and others},
  booktitle=ICCV,
  year={2025}
}

@inproceedings{zielonka2025synthetic,
  title={Synthetic prior for few-shot drivable head avatar inversion},
  author={Zielonka, Wojciech and Garbin, Stephan J and Lattas, Alexandros and Kopanas, George and Gotardo, Paulo and Beeler, Thabo and Thies, Justus and Bolkart, Timo},
  booktitle=CVPR,
  year={2025}
}

@inproceedings{zhou2025zero,
  title={Zero-1-to-A: Zero-Shot One Image to Animatable Head Avatars Using Video Diffusion},
  author={Zhou, Zhenglin and Ma, Fan and Fan, Hehe and Chua, Tat-Seng},
  booktitle=CVPR,
  year={2025}
}

@article{kirschstein2025flexavatar,
  title={FlexAvatar: Learning Complete 3D Head Avatars with Partial Supervision},
  author={Kirschstein, Tobias and Giebenhain, Simon and Nie{\ss}ner, Matthias},
  journal={arXiv preprint arXiv:2512.15599},
  year={2025}
}

@inproceedings{martinez2024codec,
  title={Codec avatar studio: Paired human captures for complete, driveable, and generalizable avatars},
  author={Martinez, Julieta and Kim, Emily and Romero, Javier and Bagautdinov, Timur and Saito, Shunsuke and Yu, Shoou-I and Anderson, Stuart and Zollh{\"o}fer, Michael and Wang, Te-Li and Bai, Shaojie and others},
  booktitle=NIPS,
  year={2024}
}

@inproceedings{ji2026fastgha,
  title={FastGHA: Generalized Few-Shot 3D Gaussian Head Avatars with Real-Time Animation},
  author={Ji, Xinya and Weiss, Sebastian and Kansy, Manuel and Naruniec, Jacek and Cao, Xun and Solenthaler, Barbara and Bradley, Derek},
  booktitle=ICLR,
  year={2026}
}

@inproceedings{kirschstein2025avat3r,
  title={Avat3r: Large animatable gaussian reconstruction model for high-fidelity 3d head avatars},
  author={Kirschstein, Tobias and Romero, Javier and Sevastopolsky, Artem and Nie{\ss}ner, Matthias and Saito, Shunsuke},
  booktitle=ICCV,
  year={2025}
}

@article{kirschstein2023nersemble,
  title={Nersemble: Multi-view radiance field reconstruction of human heads},
  author={Kirschstein, Tobias and Qian, Shenhan and Giebenhain, Simon and Walter, Tim and Nie{\ss}ner, Matthias},
  journal=TOG,
  volume={42},
  number={4},
  pages={1--14},
  year={2023},
  publisher={ACM New York, NY, USA}
}

@article{mantiuk2021fovvideovdp,
  title={Fovvideovdp: A visible difference predictor for wide field-of-view video},
  author={Mantiuk, Rafa{\l} K and Denes, Gyorgy and Chapiro, Alexandre and Kaplanyan, Anton and Rufo, Gizem and Bachy, Romain and Lian, Trisha and Patney, Anjul},
  journal=TOG,
  volume={40},
  number={4},
  pages={1--19},
  year={2021},
  publisher={ACM New York, NY, USA}
}
}

%%%%%%%%%%%%%%%%%%%%%%%%%%%%%%%%%%%%%%%%%%%%%%%%%%%%%%%%%%%%

\newpage
\appendix

% \title{\centering{-- \emph{Supplementary Material} -- }\\
% One-Shot Feed-Forward 360$^{\circ}$ Animatable Avatar via Inpainted UV-Space Gaussian Modeling}

% \maketitle
% \appendix
\section{Preliminary}
\label{sec:preliminary}

\noindent\textbf{3D Gaussian Splatting} (3DGS)~\cite{kerbl20233d} has gained recognition as a promising 3D representation, representing scenes with a collection of 3D Gaussians. Each Gaussian primitive has a series of attributes: the position $\mathbf{p} \in \mathbb{R}^3$, the scaling vector $\mathbf{s} \in \mathbb{R}^3$, the quaternion $\mathbf{q} \in \mathbb{R}^4$ representing rotation, the opacity $o \in \mathbb{R}$, and the color $\mathbf{c} \in \mathbb{R}^3$. The Gaussian is then defined as $\mathbf{G}(\mathbf{x}) = e^{-\frac{1}{2}(\mathbf{x}-\mathbf{p})^T \mathbf{\Sigma}^{-1} (\mathbf{x}-\mathbf{p})}$, where $\mathbf{\Sigma} \in \mathbb{R}^{3\times 3}$ is the covariance matrix derived from $\mathbf{s}$ and $\mathbf{q}$. Using a fast tile-based rasterizer, 3DGS enables real-time rendering.

\noindent\textbf{FLAME}~\cite{li2017learning} is a parametric face model that represents human face deformation by a combination of blendshapes and linear blend skinning (LBS), with parameters describing shape $\boldsymbol{\beta}$, pose $\boldsymbol{\theta}$, and expression $\boldsymbol{\psi}$. To deal with the deformation that LBS cannot model, additional blendshapes are applied to the template mesh vertices $\overline{\mathbf{V}}$ as:
\begin{equation}
\mathbf{V}_P=\overline{\mathbf{V}} + B_S(\boldsymbol{\beta};\mathcal{S})+ B_P(\boldsymbol{\theta};\mathcal{P})+ B_e(\boldsymbol{\psi};\mathcal{E}),
\end{equation}
where $B_S(\boldsymbol{\beta};\mathcal{S})$, $B_P(\boldsymbol{\theta};\mathcal{P})$ and $B_e(\boldsymbol{\psi};\mathcal{E})$ represent shape, pose and expression blendshapes, respectively. Then, a standard skinning function $W$ is utilized to produce the position of the deformed vertices $\mathbf{V}$:
\begin{equation}
\mathbf{V}=W(\mathbf{V}_P, \mathbf{J}(\boldsymbol{\beta}), \boldsymbol{\theta}, \mathcal{W}),
\end{equation}
where $\mathbf{J}$ is the joint locations, and $\mathcal{W}$ is the blendweights.

% \noindent\textbf{3D GAN and Inversion.}

\section{Implementation Details}
% \noindent\textbf{.}
We use the FLAME model~\cite{li2017learning} with the mouth cavity covered following~\cite{xiang2024flashavatar} as the parametric head model. We generate UV Gaussian attribute maps at $K^2 = 256^2$ resolution. To sample Gaussians from the UV Gaussian attribute maps, we use uniform grid sampling with a size of $256\times 256$, sampling one Gaussian from each UV map texel, except for the hair region, where we use the uniform grid of size $1024\times 128$ to make Gaussians cover the back of the head more completely. We also sample the UV coordinates corresponding to the FLAME vertices. The sampling results in about 78K Gaussians. We extract $N=4$ scales of the source image features for local UV feature extraction. In the symmetric UV-space inpainting module, We use 2 layers of the transformer block with convolution layers instead of linear layers to maintain the spatial structure of the UV map. The local window size $w$ is set to 7.

During training, we randomly sample 2 inverted novel views in 3D reconstruction mode, with a mode-switch probability of $0.5$. One pseudo view has 0 elevation and azimuth uniformly sampled from $[0,360^\circ)$. The other is sampled from normal distributions azimuth $\sim \mathcal{N}(0,(180^\circ)^2)$ and elevation $\sim \mathcal{N}(0,(37^\circ) ^2)$. We use the Adam optimizer with a learning rate of $1\times 10^{-4}$. The model is trained with a batch size of 4 for 37500 iterations on one NVIDIA H800 GPU. The loss weights are set as $\lambda_{3d}=50$, $\lambda_{eye}=5$, $\lambda_{pos}=1$, $\lambda_{shape}=1$, $\lambda_{shape}^{tv}=10$, $\lambda_{id}=0.25$, $\lambda^{1}=0.5$, and the threshold is set to $\epsilon=0.1$. We add $\mathcal{L}_{3d}$, $\mathcal{L}_{eye}$, $\mathcal{L}_{shape}^{tv}$ and $\mathcal{L}_{id}$ after training for 7500 iterations, and increase $\lambda_{3d}$ to $100$ after training for 25000 iterations.

\begin{figure*}[tb]
    \centering
  \includegraphics[width=1\linewidth]{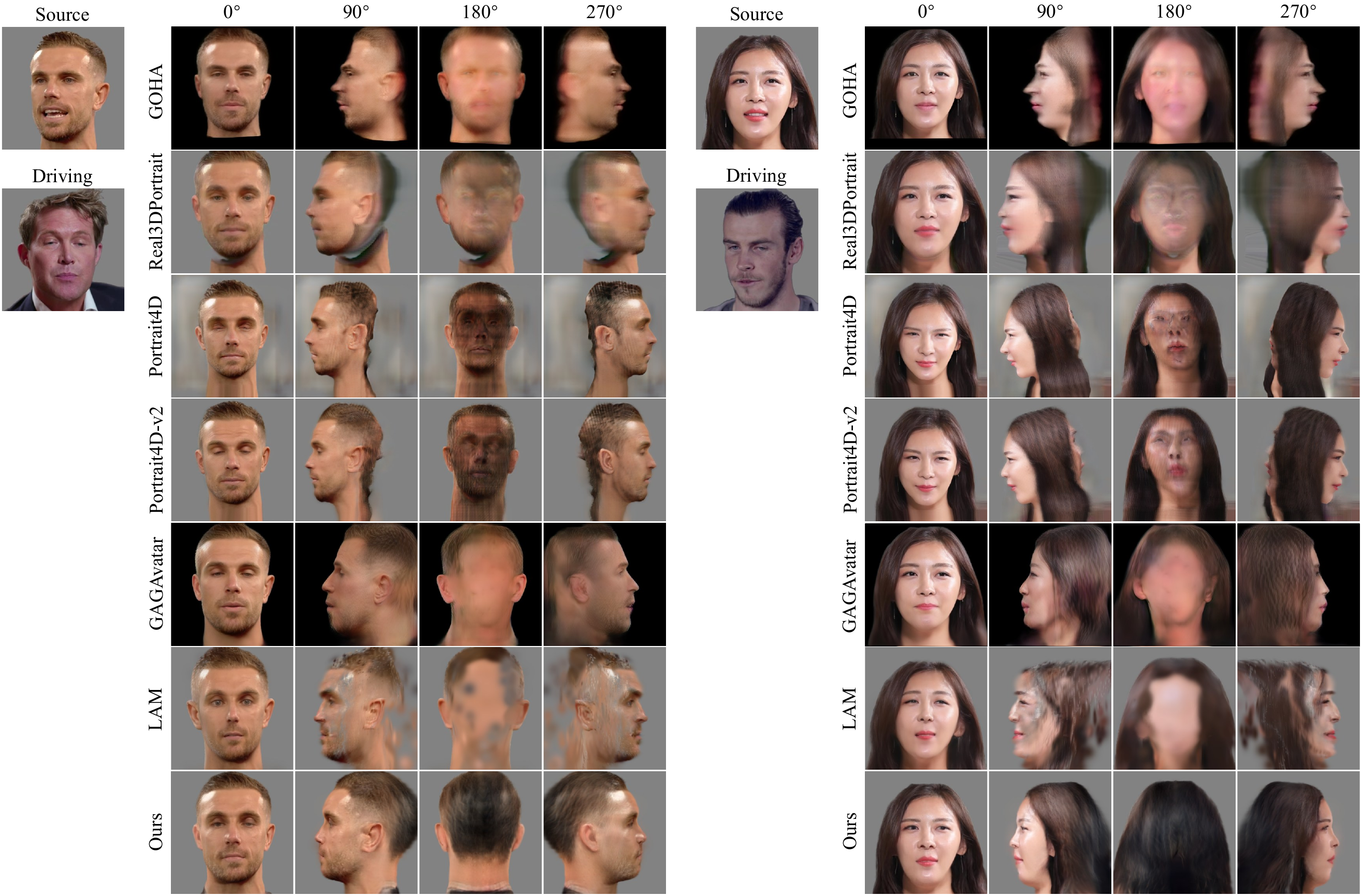}
  \vspace{-14pt}

  \caption{Qualitative comparison with state-of-the-art 3D-based methods (\ie, GOHA~\cite{li2023generalizable}, Real3DPortrait~\cite{ye2024realdportrait}, Portrait4D~\cite{deng2024portrait4d}, Portrait4D-v2~\cite{deng2024portrait4d_eccv}, GAGAvatar~\cite{chu2024generalizable} and LAM~\cite{he2025lam}) for 360$^\circ$ rendering on the VFHQ~\cite{xie2022vfhq} dataset. Our method renders plausible novel views across 360$^\circ$, whereas other approaches fail at large side or back views.
  }
  \vspace{-6pt}
  \label{fig:sota_cmp_multiview}
\end{figure*}

\section{Additional Details on Experiments}

\subsection{Additional Comparison Details}
We compare our method with a group of open-source state-of-the-art methods, \ie, the 2D-based StyleHEAT~\cite{yin2022styleheat}, the 3D-based GOHA~\cite{li2023generalizable}, Real3DPortrait~\cite{ye2024realdportrait}, Portrait4D~\cite{deng2024portrait4d}, Portrait4D-v2~\cite{deng2024portrait4d_eccv}, GAGAvatar~\cite{chu2024generalizable} and LAM~\cite{he2025lam}. For each approach, we follow the official data pre-processing and animation procedures, and use the released checkpoint for evaluation. To compute the metrics, we use the processed and generated frames for each method, as they are typically aligned and share the same background color. However, Portrait4D~\cite{deng2024portrait4d} and Portrait4D-v2~\cite{deng2024portrait4d_eccv} generate misaligned images with their processed frames, so we further align these images before computing the metrics.

\subsection{Additional Experimental Results}

\begin{figure*}[tb]
    \centering
  \includegraphics[width=1\linewidth]{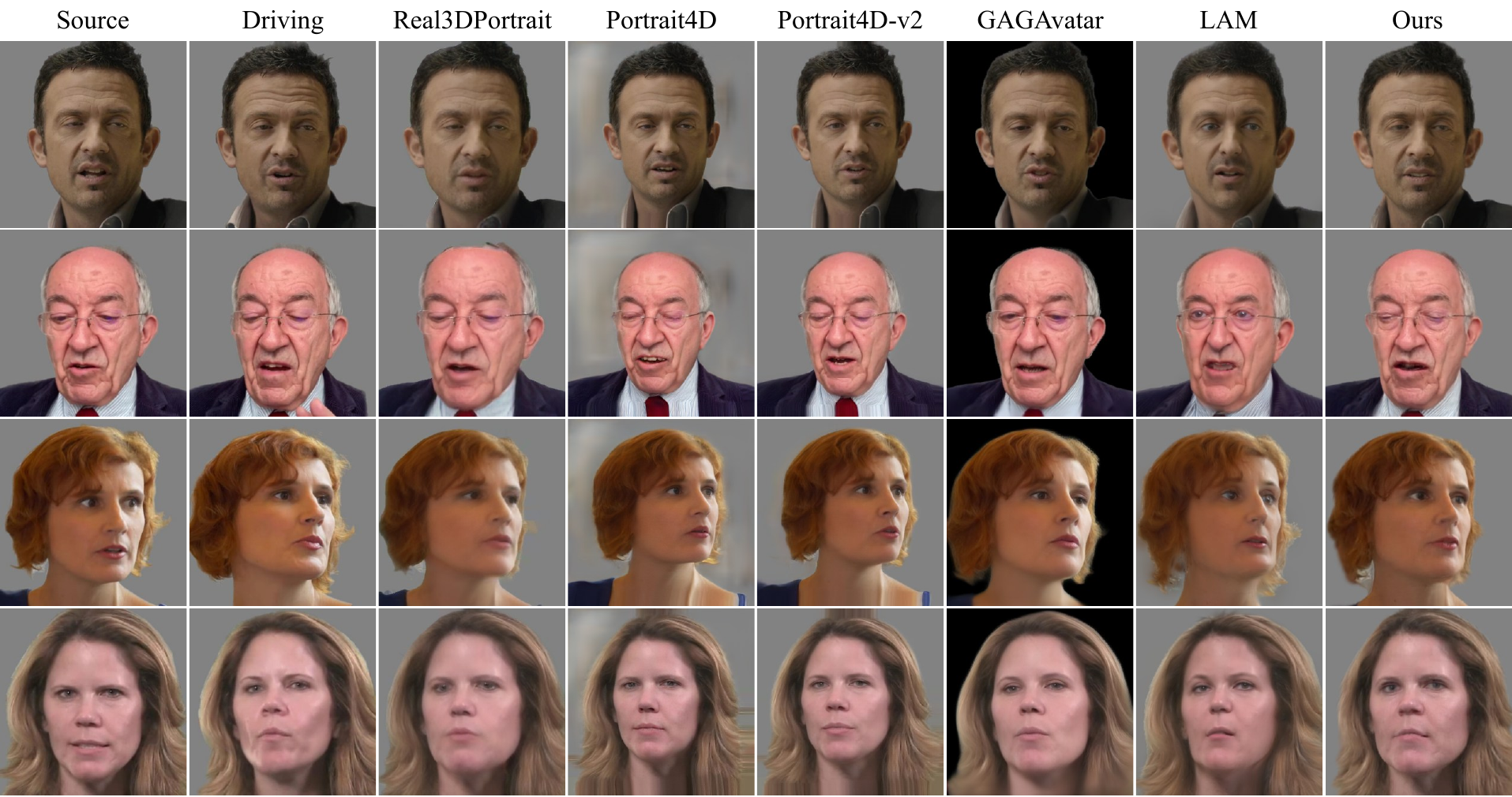}
  \vspace{-14pt}

  \caption{Qualitative comparison with state-of-the-art methods (\ie, Real3DPortrait~\cite{ye2024realdportrait}, Portrait4D~\cite{deng2024portrait4d}, Portrait4D-v2~\cite{deng2024portrait4d_eccv}, GAGAvatar~\cite{chu2024generalizable} and LAM~\cite{he2025lam}) for self-reenactment on the VFHQ~\cite{xie2022vfhq} and HDTF~\cite{zhang2021flow} datasets. Our method best preserves source appearance details while effectively mimicking the driving motion.
  }
  \vspace{-6pt}
  \label{fig:sota_cmp_sameid}
\end{figure*}

\noindent\textbf{Video Demos.}
We provide \textcolor{black}{video demos on the project page\footnote{\href{https://shaelynz.github.io/fhavatar/}{https://shaelynz.github.io/fhavatar/}}}. These video demos present results for comparison with state-of-the-art approaches in terms of avatar animation and 360$^\circ$ rendering. They demonstrate the effectiveness of our method, supporting high-quality animation and 360$^\circ$ rendering views.
%The avatar generated by our method preserves source identity the best while accurately mimicking the driving motion. It also maintains the best 3D consistency at large side or back views, whereas other approaches generally collapse.

% \subsubsection{Additional Qualitative Comparison Results}
% \noindent\textbf{Full-Head Rendering.}
\noindent\textbf{Additional 360$^\circ$ Rendering Results. }
We compare the quality of 360$^\circ$ rendering with the 3D-based methods in Fig.~\ref{fig:sota_cmp_multiview}. The avatar is animated by the driving motion before being rendered to different viewpoints across 360$^\circ$. All other methods fail at large side or back views that significantly differ from the input source view, unable to model the back of the head. Only GAGAvatar~\cite{chu2024generalizable} can roughly model the shape of the back head at side views, but the textures are coarse and contain visible artifacts. Meanwhile, due to its use of a 2D neural renderer on the rendered features, GAGAvatar cannot maintain 3D consistency of the face area under large camera pose changes. In contrast, by inpainting UV-space source local features, our method reconstructs full-head avatars with plausible details for the back of the head, improving the realism of animatable 3D head avatars.

% \noindent\textbf{Self-reenactment.}
\noindent\textbf{Self-reenactment.}
We provide qualitative comparisons of methods for self-reenactment in Fig.~\ref{fig:sota_cmp_sameid}. Our method can faithfully preserve appearance details in the source image and imitate the driving motion at the same time.

% \begin{wraptable}{r}{0.72\linewidth}
\begin{table}[tb]
\vspace{-4pt}
\caption{Favorability rate of our method in the user study.}
\vspace{-4pt}
  \begin{center}
  \resizebox{0.8\linewidth}{!}{
  \setlength{\tabcolsep}{3mm}
  \begin{tabular}{l|ccc}
  \toprule[1.2pt]
     {Compared method} & { vs. Portrait4D-v2~\cite{deng2024portrait4d_eccv}} & { vs. GAGAvatar~\cite{chu2024generalizable} }& { vs. LAM~\cite{he2025lam}} \\
    \midrule
     { Favorability rate $\uparrow$} &{ 61.5\%} &{ 56.0\%} &{ 83.5\%}   \\  
    \bottomrule[1.2pt]
  \end{tabular}
  }
  \end{center}
  % \vspace{-22pt}
  
  \vspace{-6pt}
  \label{reb:table:user_study}
\end{table}
% \end{wraptable}

\noindent\textbf{User Study for Cross-identity Reenactment.}
We conduct a user study with 20 participants to compare 10 cross-identity reenactment results from our method with recent state-of-the-art approaches in terms of appearance preservation, motion replication, and overall quality.  Tab.~\ref{reb:table:user_study} shows that users prefer our results, confirming our generation quality.
% The table above shows that users prefer our results.

\begin{table}[tb]
% \vspace{-8pt}
\caption{Quantitative comparison for identity consistency under stratified yaw angles.}
  \label{reb:table:stratified_finer}
    \vspace{-4pt}
  \begin{center}
  \resizebox{1\linewidth}{!}{
  \setlength{\tabcolsep}{3mm}
  % \Large
  \begin{tabular}{l|c|c|c|c|c|c}
  \toprule[1.2pt]
         
         % \multirow{3}{*}{{ Method}}  & \multicolumn{4}{c|}{Ava-256} & \multicolumn{3}{c}{SphereHead generated}\\
         % \cmidrule{2-8}
        \multirow{2}{*}{{ Method}} & \multicolumn{1}{c|}{ { $0\sim 30^\circ$}} & \multicolumn{1}{c|}{ { $30^\circ \sim 60^\circ$}} & \multicolumn{1}{c|}{ \textcolor{black}{ $60^\circ \sim 90^\circ$}} & $90^\circ \sim 120^\circ$   & $120^\circ \sim 150^\circ$   & $150^\circ \sim 180^\circ$  \\
         % \cmidrule{2-8}
      %& LPIPS $\downarrow$
      & CSIM $\uparrow$ %& LPIPS $\downarrow$
      & CSIM $\uparrow$ %& LPIPS $\downarrow$
      & CSIM $\uparrow$ & LPIPS $\downarrow$  & LPIPS $\downarrow$  & LPIPS $\downarrow$  \\
        \midrule
         
       Portrait4D-v2~\cite{deng2024portrait4d_eccv}  %& 0.3037 
       & 0.8189 %& 0.3977 
       & 0.5564 %& 0.3671 
       & 0.4577 & 0.5282 & 0.5042 & 0.5126  \\ 
        GAGAvatar~\cite{chu2024generalizable}  %& \textbf{0.2364}
        & \textbf{0.8444} %& \textbf{0.3505} 
        & 0.5913 %& \textbf{0.3599} 
        & 0.4157 & 0.5190 & 0.5339 & 0.5348 \\

       Ours %& 0.2650 
       & 0.8242 %& 0.3765 
       & \textbf{0.6055} %& 0.3622 
       & \textbf{0.5190} & \textbf{0.4286} & \textbf{0.4335} & \textbf{0.4242} \\

    \bottomrule[1.2pt]
  \end{tabular}
  }
  \end{center}
  % \vspace{-25pt}

  \vspace{-6pt}
\end{table}

% \vspace{-12pt}
\begin{figure}[tb]
    \begin{center}
  \includegraphics[width=0.8\linewidth]{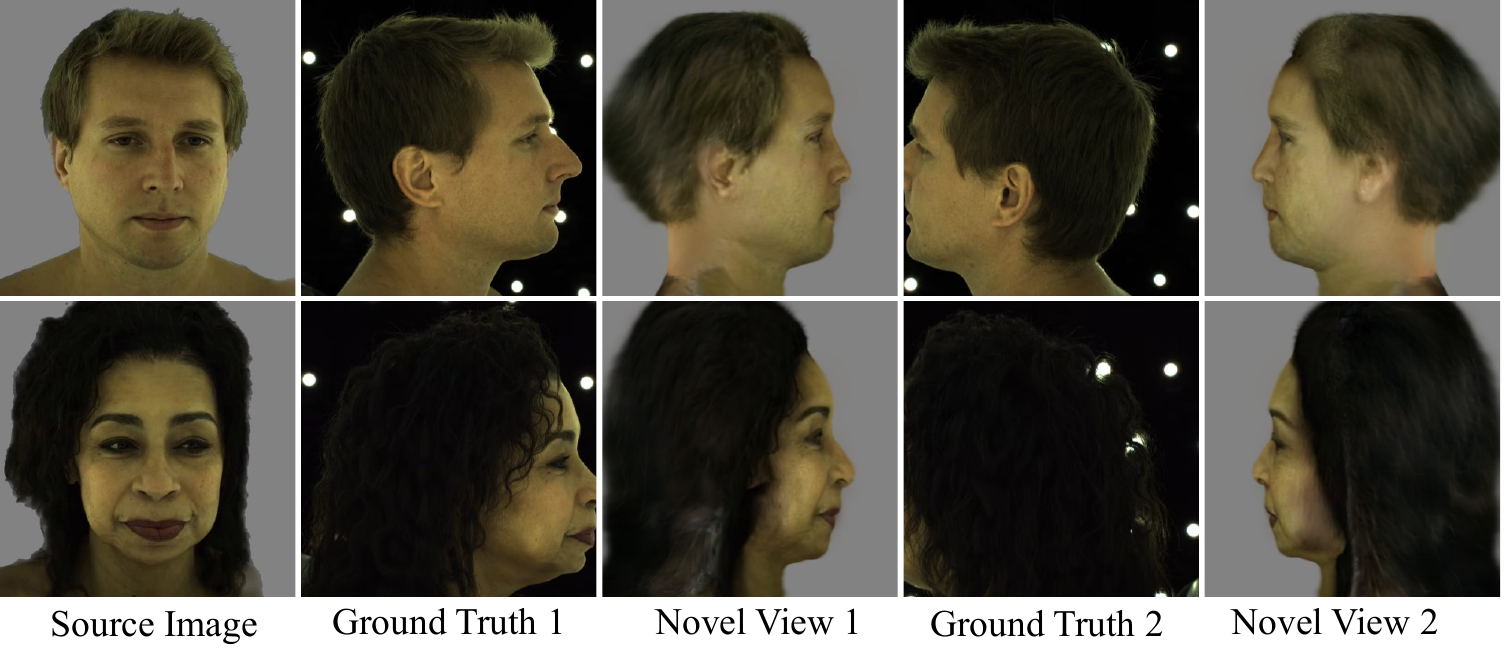}
  \end{center} 
    \vspace{-8pt}
    \caption{Qualitative analysis of identity consistency at large poses.}
    \label{reb:fig:large_pose_id}
    % \vspace{-10pt}
\end{figure}

\noindent\textbf{Analysis of Identity Consistency under Different Poses.}
We present a stratified quantitative evaluation for identity consistency on yaw ranges relative to the front view (0$\sim$180$^\circ$ covering the full 360$^\circ$ range). Specifically, we evaluate on Ava-256~\cite{martinez2024codec} for yaw angles of $0$ to $90^\circ$ and on images generated by SphereHead~\cite{li2024spherehead} for yaw angles of $90^\circ+$ in Tab.~\ref{reb:table:stratified_finer}. %{reb:table:stratified}.
% We use LPIPS for $90^\circ +$ evaluation as no faces can be detected.
For yaw angles beyond $90^\circ$, we use LPIPS as CSIM becomes unreliable when faces are not detected.
For Ava-256, we randomly select 8 identities with 7 views. For SphereHead, we generate 100 identities and evaluate at 2 random yaw angles in each stratified yaw angle range. The most frontal views serve as input. Our method excels at preserving identity for poses with yaw angles of $30^\circ+$.
We also verify identity consistency at large poses with ground-truths from Ava-256 in Fig.~\ref{reb:fig:large_pose_id}. 
Being limited by the GAN prior and the ambiguity of a single frontal view, the generated novel views lack hair details, but they still preserve convincing back heads and the source identity.

% \color{red}
\color{black}
\noindent\textbf{Temporal Consistency for Animation.}
We measure temporal consistency of the generated videos 
on VFHQ~\cite{xie2022vfhq} using JOD~\cite{mantiuk2021fovvideovdp} following Avat3r~\cite{kirschstein2025avat3r}. Tab.~\ref{reb:table:temporal} shows that our method achieves the best consistency. 

\begin{table}[tb]
\caption{Quantitative comparison on temporal consistency.}
\label{reb:table:temporal}
\vspace{-2pt}
  \begin{center}
  \resizebox{0.6\linewidth}{!}{
  \Large
  \begin{tabular}{l|cccc}
  \toprule[2pt]
     { Method} & { Portrait4D-v2~\cite{deng2024portrait4d_eccv} } & { GAGAvatar~\cite{chu2024generalizable} }& { LAM~\cite{he2025lam}} & { Ours} \\
    \midrule
     { JOD $\uparrow$} &{ 4.5067} &{ 5.0585} &{ 4.6943}  &{ \textbf{5.3410}} \\  
    \bottomrule[2pt]
  \end{tabular}
  }
  \end{center}
  % \vspace{-22pt}
  % \caption{Quantitative comparison on temporal consistency.}
  % \vspace{-13pt}
\end{table}

\begin{figure*}[tb]
    % \vspace{-18pt}
    \centering
  \includegraphics[width=1\linewidth]{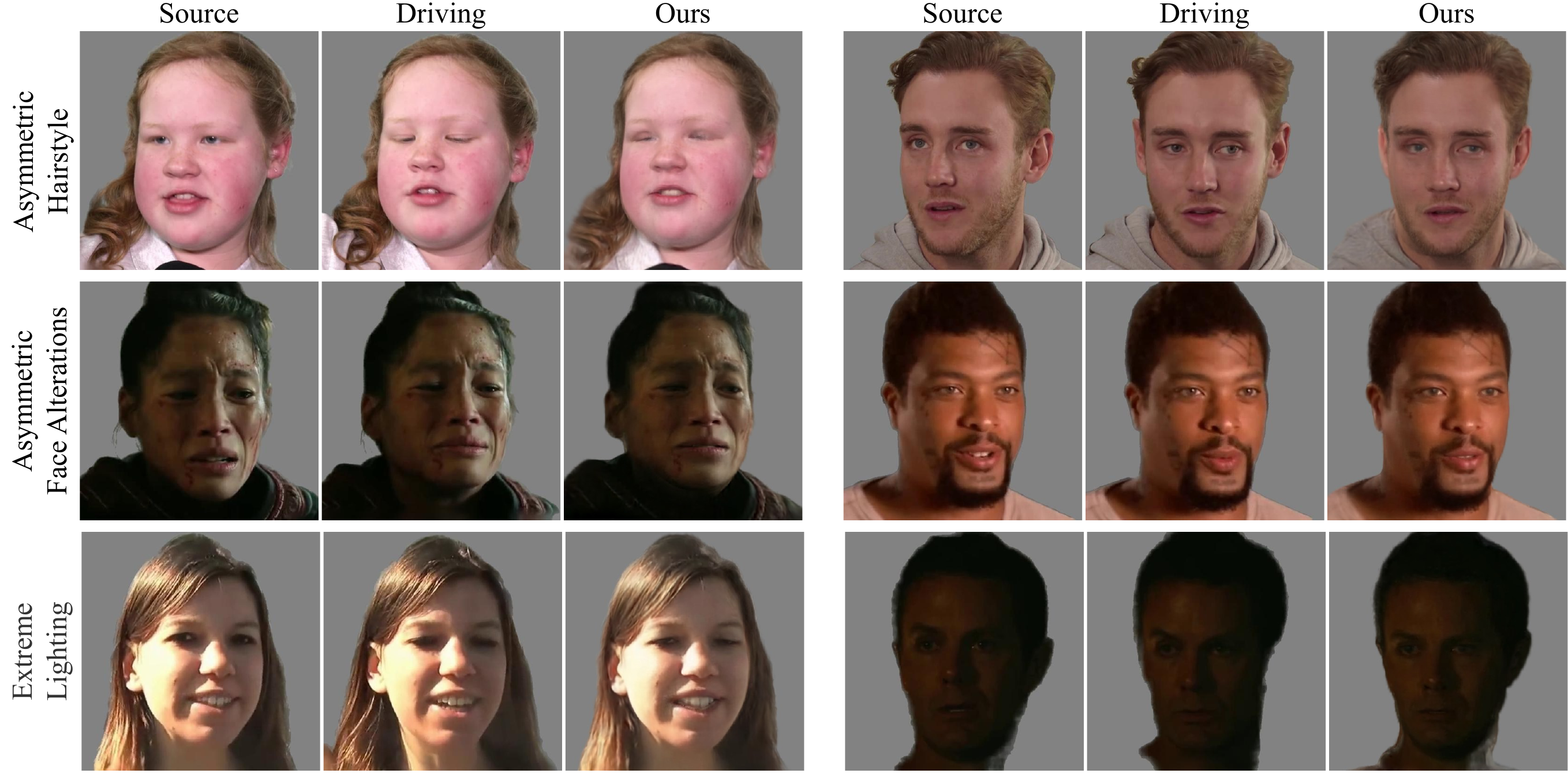}
  \vspace{-18pt}

  \caption{Qualitative results for robustness analysis. Our method is robust under asymmetric hairstyles, asymmetric face alterations and extreme lighting conditions.}
  \vspace{-8pt}
  \label{fig:robust}
\end{figure*}

\color{black}
% \noindent\textbf{Robustness Analysis.}
\noindent\textbf{Robustness Analysis.}
We perform a robustness analysis of our method under asymmetric hairstyles, asymmetric face alterations and extreme lighting conditions in Fig.~\ref{fig:robust}. As our method mainly utilizes symmetry priors for the generation of occluded areas, it can preserve asymmetric hairstyles or face alterations that are visible in the input image. Our method is also robust under extreme lighting conditions.

% \color{red}

\begin{table}[!b]
\caption{More quantitative ablation study on 3D GAN priors.}
  \label{table:ablation_3dprior_mead}
  % \vspace{-6pt}
  \centering
  \resizebox{0.99\linewidth}{!}{
  \Large
  \begin{tabular}{l|c|c|ccccccc} 
  \toprule[2pt]
    Method  
             & + Real multi-view training data
             & 360$^\circ$ view synthesis
             & PSNR $\uparrow$  & SSIM $\uparrow$   & LPIPS $\downarrow$ & CSIM $\uparrow$  & AKD $\downarrow$ & AED $\downarrow$ & APD $\downarrow$
             \\
    \midrule
       w/o pseudo multi-view training data
    & \ding{56}
    & \ding{56}
    & \textit{17.82} & \textit{0.7504} & \textit{0.3356} & \textit{0.6843} & \textit{4.930} & \textit{0.4958} & \textit{2.5626}
    \\
    \midrule
   w/o global UV maps
    & \ding{56}
    & \ding{52}
    & \textbf{17.20} & 0.7471 & 0.3434 & 0.6701 & 4.944 & 0.5101 & 2.1511
    \\
    Ours 
    & \ding{56}
    & \ding{52}
    & \textbf{17.20} & \textbf{0.7495} & \textbf{0.3402} & \textbf{0.6765} & \textbf{4.844} & \textbf{0.5037} & \textbf{1.9484}
    \\
    % \midrule[1pt]
    % w/o pseudo multi-view training data 
    % & \ding{52}
    % & \ding{56}
    % & \textit{17.95} & \textit{0.7575} & \textit{0.3288} & \textit{0.6749} & \textit{4.908} & \textit{0.5141} & \textit{2.1513}   \\
    
    \midrule
    % w/o global UV maps
    % & \ding{52}
    % & \ding{52}
    % & 17.49 & 0.7533 & 0.3370 & 0.6722 & 4.942 & 0.5054 & 2.2119

    % \\ 
    
    Ours
    & \ding{52}
    & \ding{52}
    & \textit{17.51} & \textit{0.7558} & \textit{0.3313} & \textit{0.6742} & \textit{4.876} & \textit{0.4896} & \textit{2.1592}
    
    \\ 

    \bottomrule[2pt]
  \end{tabular}
  }
  % \vspace{-20pt}

  % \vspace{-12pt}
\end{table}

% \begin{wrapfigure}{r}{0.38\linewidth}
\begin{figure}[!b]
    % \vspace{-25pt}
    \begin{center}
    \includegraphics[width=1\linewidth]{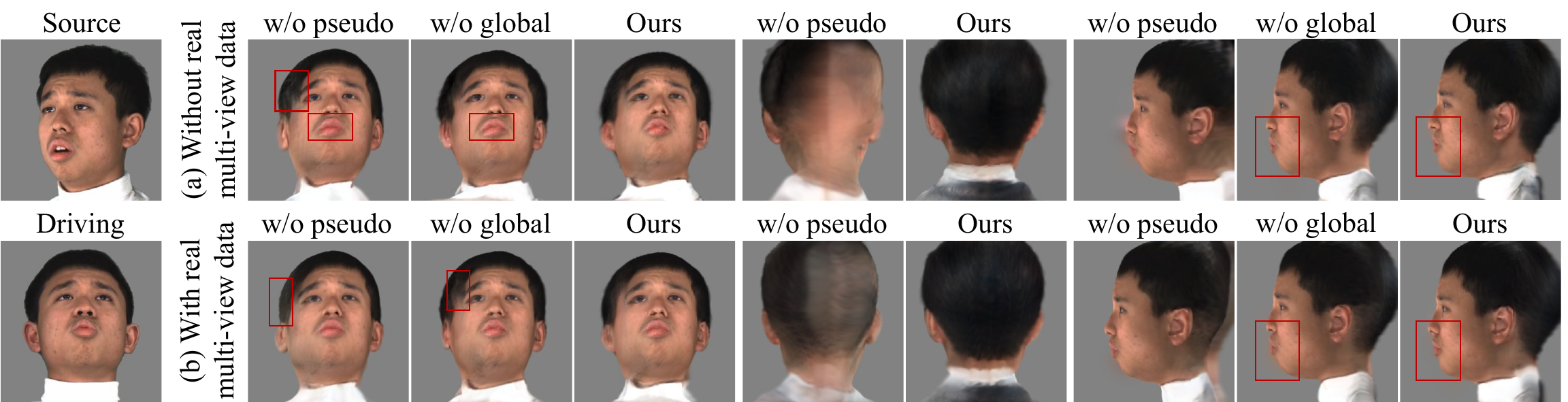}
  \end{center} 
    \vspace{-6pt}
    \caption{More qualitative ablation study on 3D GAN priors.}
    \label{reb:fig:ablation_3dprior_mead}
    % \vspace{-12pt}
\end{figure}
% \end{wrapfigure}

\color{black}
\noindent\textbf{More Ablation Study on 3D GAN Priors.}
To isolate the effects of global UV feature maps and pseudo multi-view data from 3D GAN priors, ablations on multi-view MEAD~\cite{wang2020mead} are shown in Tab.~\ref{table:ablation_3dprior_mead} and Fig.~\ref{reb:fig:ablation_3dprior_mead}. 
Global UV feature maps offer global face structures, resulting in consistent improvement, especially under occlusion (\eg, lips in Fig.~\ref{reb:fig:ablation_3dprior_mead}). Pseudo multi-view data supplies essential 360$^\circ$ supervision; without it, significant side/back artifacts occur. Moreover, real multi-view data from NeRSemble~\cite{kirschstein2023nersemble} enhances animation quality, further improving side views when combined with 3D GAN priors.

% \begin{figure}[ht]
% \centering
% \begin{minipage}[t]{0.39\linewidth}
%     \vspace{-9pt}
%         \centering
%     \begin{table}[ht]
%       \begin{center}
%       \resizebox{1\linewidth}{!}{
%       \Large
%       \begin{tabular}{l|cc}
%       \toprule[2pt]
%        Method & PSNR $\uparrow$ & SSIM $\uparrow$\\
%         \midrule
%         Ours & 17.53 & 0.6051 \\
%         + Hair Gaussian Rescaling & \textbf{17.57} & \textbf{0.6052} \\
%         \bottomrule[2pt]
%       \end{tabular}
%       }
%       \end{center}
%       \caption{Quantitative comparison on back head quality.}
%       \label{reb:table:back_head}
%       \end{table}
%     % \vspace{-23pt}
% \end{minipage}
% \hfill
% \begin{minipage}[t]{0.5\linewidth}
%         \begin{figure}[ht]
%         \vspace{-12pt}
%         \centering
%         % \includegraphics[width=1\linewidth]{results/exp_qualitative_multiview_more_addrescale.pdf}
%         \includegraphics[width=1\linewidth]{results/addrescale_back_source.pdf}
%         \caption{Qualitative comparison on blurriness at extreme views.}
%         \label{reb:fig:more_360}
%         \end{figure}
%     % \vspace{-23pt}
% \end{minipage}

% \vspace{-23pt}
% \end{figure}

\begin{figure}[!t]
    % \vspace{-24pt}
    \begin{center}
  \includegraphics[width=1\linewidth]{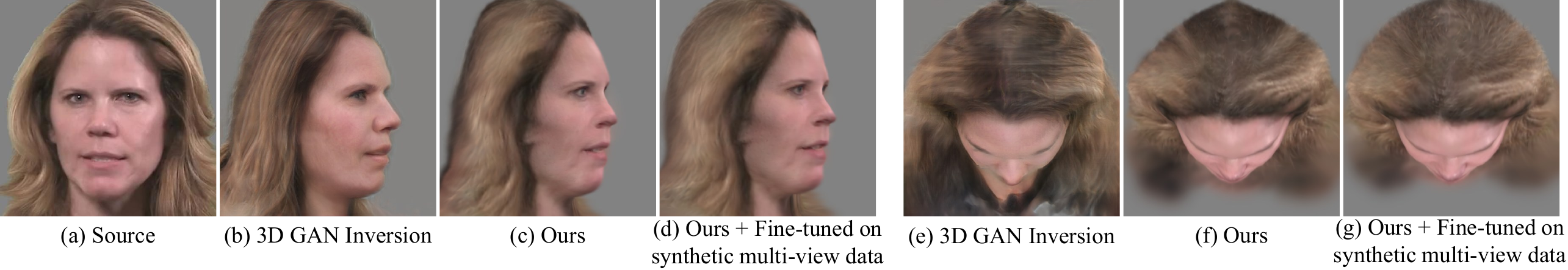}
  \end{center} 
    \vspace{-6pt}
    \caption{Qualitative results of head shapes at extreme views. Our results can be further improved by fine-tuning on synthetic multi-view data.}
    \label{reb:fig:headshape}
    % \vspace{-23pt}
\end{figure}

\begin{figure}[!t]
  % \vspace{-20pt}
  \centering
  
  % ------ 左边的表格 ------
  \begin{minipage}[c]{0.4\textwidth}
  \vspace{-15pt}
    \centering
    \captionof{table}{Quantitative comparison on back head quality.}
    \label{reb:table:back_head}
    \resizebox{0.99\linewidth}{!}{
    \begin{tabular}{l|cc}
      \toprule[1.2pt]
       Method & PSNR $\uparrow$ & SSIM $\uparrow$\\
        \midrule
        Ours & 17.53 & 0.6051 \\
        + Hair Gaussian Rescaling & \textbf{17.57} & \textbf{0.6052} \\
        \bottomrule[1.2pt]
      \end{tabular}
      }
  \end{minipage}
  \hfill % 弹性间距，让左右两个 minipage 自动撑开
  % ------ 右边的图片 ------
  \begin{minipage}[c]{0.55\textwidth}
  % \vspace{-12pt}
    \centering
    % 替换为你的图片路径，[width=\textwidth] 表示撑满当前 minipage 的宽度
    \includegraphics[width=1\textwidth]{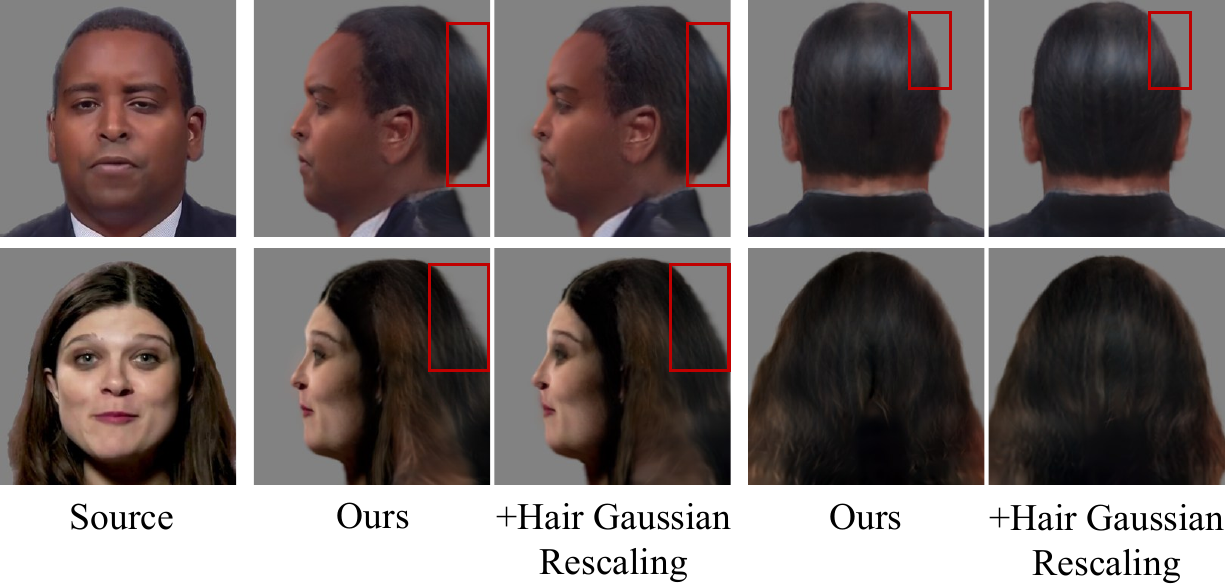}
    \vspace{-14pt}
    \captionof{figure}{Qualitative comparison on blurriness at extreme views.}
    \label{reb:fig:blurriness_extremeview}
  \end{minipage}

\end{figure}

\noindent\textbf{Further Improving Visual Quality at Extreme Views.}
Fig.~\ref{fig:sota_cmp_multiview} confirms the strong visual quality of our method at extreme views. We further show that visual results can be improved by adjusting two factors of our method: 
(i) Using pseudo multi-view images from 3D GAN inversion as 3D supervision constrains head geometry, leading to a slightly pointed back head, as shown in Fig.~\ref{reb:fig:headshape}. Fine-tuning on multi-view data generated by SphereHead~\cite{li2024spherehead} for 500 iterations alleviates this artifact.
(ii) Overlarge Gaussians in the back head region cause blurriness. We mitigate this by rescaling Gaussians in the hair region by 0.5, which improves back head clarity. This leads to better reconstruction on back-view images generated by SphereHead in Tab.~\ref{reb:table:back_head} and visibly sharper side/back views in Fig.~\ref{reb:fig:blurriness_extremeview}. 
\color{black}

% \begin{figure}[htbp]
% \floatsetup{style=plain,subcapbesideposition=top}
% \ffigbox[\FBwidth]{
%   \includegraphics[width=0.5\linewidth]{results/addrescale_back_source.pdf}
% }{
%   \caption{Qualitative comparison on blurriness at extreme views.}
% }
% \hfil
% \ttabbox[\FBwidth]{
%   \begin{tabular}{ccc}
%     A & B & C \\
%     1 & 2 & 3
%   \end{tabular}
% }{
%    \caption{Quantitative comparison on back head quality.}
% }
% \end{figure}

% \begin{table}[tb]
% \caption{Quantitative comparison for identity consistency under stratified yaw angles.}
% \vspace{-18pt}
%   \begin{center}
%   \resizebox{0.8\linewidth}{!}{
%   \setlength{\tabcolsep}{3mm}
%   \begin{tabular}{l|c|c|c|c}
%   \toprule[1.2pt]

%         \multirow{2}{*}{{Method}} & \multicolumn{1}{c|}{ { $0\sim 30^\circ$}} & \multicolumn{1}{c|}{ { $30^\circ \sim 60^\circ$}} & \multicolumn{1}{c|}{ \textcolor{black}{ $60^\circ \sim 90^\circ$}} & $90^\circ+$  \\

%       & CSIM $\uparrow$ %& LPIPS $\downarrow$
%       & CSIM $\uparrow$ %& LPIPS $\downarrow$
%       & CSIM $\uparrow$ & LPIPS $\downarrow$  \\
%         \midrule
 
%        Portrait4D-v2~\cite{deng2024portrait4d_eccv} %& 0.3037 
%        & 0.8189 %& 0.3977 
%        & 0.5564 %& 0.3671 
%        & 0.4577 & 0.5090 \\ 

%         GAGAvatar~\cite{chu2024generalizable} %& \textbf{0.2364}
%         & \textbf{0.8444} %& \textbf{0.3505} 
%         & 0.5913 %& \textbf{0.3599} 
%         & 0.4157 & 0.5335 \\       

%        Ours %& 0.2650 
%        & 0.8242 %& 0.3765 
%        & \textbf{0.6055} %& 0.3622 
%        & \textbf{0.5190} & \textbf{0.4299} \\

%     \bottomrule[1.2pt]
%   \end{tabular}
%   }
%   \end{center}
%   \vspace{-10pt}
%   \label{reb:table:stratified}
% \end{table}

\begin{figure}[!b]
    % \vspace{-25pt}
    \begin{center}
    \includegraphics[width=1\linewidth]{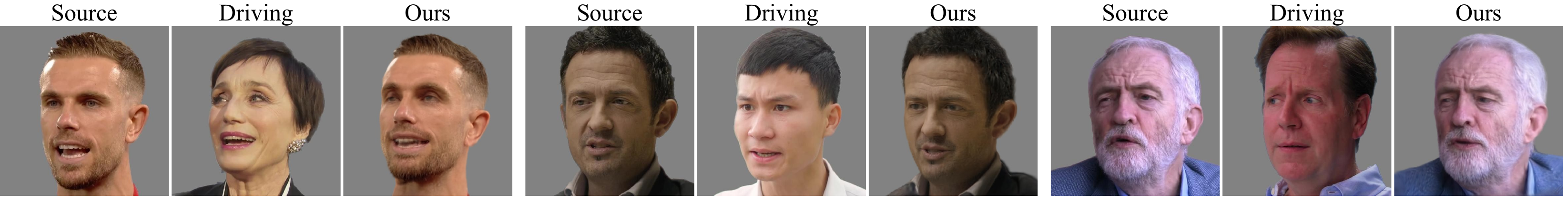}
  \end{center} 
    \vspace{-6pt}
    \caption{Qualitative results on challenging motions.}
    \label{reb:fig:challenging_motion}
    % \vspace{-18pt}
\end{figure}

% \color{red}
\color{black}
\noindent\textbf{Additional Motion Modeling and Evaluation.}
Our method uses FLAME for animation, which primarily captures motion-related deformation but fails to model motion-related appearance changes, such as wrinkles. As shown in Tab.~\ref{table:motion_module}, incorporating a lightweight motion module for expression-aware UV appearance further improves animation. 
Fig.~\ref{reb:fig:challenging_motion} shows our method can handle challenging expressions, such as large smiles and frowns.

\begin{table}[!b]
\caption{Quantitative results on additional motion modeling}
\vspace{-4pt}
  \label{table:motion_module}
\begin{center}
  \resizebox{1\linewidth}{!}{ %
  % \Large
  \begin{tabular}{l|ccccccc|ccc} %l}
  \toprule[1.2pt]
    \multirow{2}{*}{Method}  &  \multicolumn{7}{c|}{Self-reenactment}  &  \multicolumn{3}{c}{Cross-identity Reenactment} \\
    %\cline{2-7} \cline{8-13}
    
             & PSNR $\uparrow$  & SSIM $\uparrow$   & LPIPS $\downarrow$ & CSIM $\uparrow$  & AKD $\downarrow$ & AED $\downarrow$ & %APD $\downarrow$
             APD $\downarrow$
             & CSIM $\uparrow$  & AED $\downarrow$ & APD $\downarrow$\\
    \midrule    
    Ours 
    & {23.24} & {0.7995}  & {0.2384} &  0.8012  &  {2.798}  & 0.3634  & %\underline{0.0139}
    {0.8660}
    & {0.6757} & 0.7103  & 2.3661
    \\ 
    Ours + motion module 
    & \textbf{23.25} & \textbf{0.8014} &\textbf{ 0.2375} & \textbf{0.8138} & \textbf{2.789} & \textbf{0.3562} & \textbf{0.8410} 
    & \textbf{0.6965} & \textbf{0.7060} & \textbf{2.2667}
    \\
    \bottomrule[1.2pt]
  \end{tabular}
  }
  \end{center}
  % \vspace{-20pt}
\end{table}

\color{black}

% \begin{wraptable}{r}{0.5\linewidth}
\begin{table}[h]
\caption{Quantitative comparison with 3D GAN inversion~\cite{bilecen2024dual} for one-shot static head reconstruction. }
\vspace{-2pt}
  \begin{center}
  \resizebox{0.5\linewidth}{!}{
  \setlength{\tabcolsep}{3mm}
  \begin{tabular}{l|cc}
  \toprule[1.2pt]
     Method & LPIPS $\downarrow$ & CSIM $\uparrow$ \\
    \midrule
     3D GAN inversion~\cite{bilecen2024dual} & 0.2988 & 0.5044   \\  
     Ours & \textbf{0.2545} & \textbf{0.8009}  \\  
    \bottomrule[1.2pt]
  \end{tabular}
  }
  \end{center}
  % \vspace{-18pt}
  \label{reb:table:panohead_comparison}
\end{table}
% \end{wraptable}

\noindent\textbf{Comparison with 3D GAN Inversion.}
We compare with the 3D GAN inversion method~\cite{bilecen2024dual} used in our framework for one-shot static head reconstruction in Tab.~\ref{reb:table:panohead_comparison} by reconstructing each frame of 17 videos from the VFHQ~\cite{xie2022vfhq} test split. Our method better preserves the input identity and appearance details, which indicates the importance of inpainting input local UV feature maps rather than simply relying on the output of 3D GAN inversion.

\begin{table}[tb]
\caption{Quantitative results of our method without using pseudo multi-view data during training. Although it achieves better animation performance, it cannot support 360$^\circ$ view synthesis.
  }
  \vspace{-2pt}
  \begin{center}
  \resizebox{0.99\linewidth}{!}{
  \setlength{\tabcolsep}{1mm}
  \begin{tabular}{l|l|ccccccc|ccc}
  \toprule[1.2pt]
        \multirow{2}{*}{{Dataset}} 
        & \multirow{2}{*}{{ Method}} & \multicolumn{7}{c|}{ { Self-reenactment}} & \multicolumn{3}{c}{ { Cross-identity Reenactment}} \\
     & & PSNR $\uparrow$& SSIM $\uparrow$& LPIPS $\downarrow$& CSIM $\uparrow$& AKD $\downarrow$& AED $\downarrow$& APD $\downarrow$ & CSIM $\uparrow$ &AED $\downarrow$ &APD $\downarrow$ \\
    \midrule
     \multirow{2}{*}{{VFHQ}} & 
      
       Ours & {23.24} & {0.7995} & {0.2384} & 0.8012 &{2.798} & 0.3634 & \textbf{0.8660} & {0.6757} & 0.7103 & \textbf{2.3661}  \\ 
      %\midrule
      % \cmidrule{2-12}
      & Ours (w/o pseudo) & \textbf{23.62} & \textbf{0.8064} & \textbf{0.2304} & \textbf{0.8051} & \textbf{2.703} & \textbf{0.3446} & 0.8779 & \textbf{0.6940} & \textbf{0.6872} & 2.5628 \\
      
     \midrule[1pt]
     \multirow{2}{*}{{HDTF}} & 
       Ours &{26.61} &{0.8642} &{0.1900} & \textbf{0.8622} & {2.287} & {0.3318} & \textbf{0.5286} & 0.8568 & 0.7347 & 1.5605  \\ 
      %\midrule
      % \cmidrule{2-12}
      & Ours (w/o pseudo) & \textbf{26.93} & \textbf{0.8689} & \textbf{0.1860} & 0.8612 & \textbf{2.252} & \textbf{0.3258 }& 0.5643 & \textbf{0.8601} & \textbf{0.7262} & \textbf{1.5587 }\\
    \bottomrule[1.2pt]
  \end{tabular}
  }
  \end{center}
  % \vspace{-20pt}
  \label{reb:table:pseudo_multi_view_data}
\end{table}

\noindent\textbf{Fairness of Using Pseudo Multi-View Data.} 
We use pseudo multi-view data for novel-view supervision during training, following Real3DPortrait~\cite{ye2024realdportrait} and Portrait4D-v2~\cite{deng2024portrait4d_eccv}. Specifically, we use pseudo multi-view data from 3D full-head GAN inversion~\cite{an2023panohead, bilecen2024dual} to realize 360$^\circ$ view synthesis. 
To make a fair comparison when such pseudo multi-view data is used, we also present results
without using pseudo multi-view data in our method, labeled ``Ours (w/o pseudo)" in Tab.~\ref{reb:table:pseudo_multi_view_data}. It generally yields better performance, likely because it is not affected by the inverted pseudo multi-view data, which fails to preserve input image details and identity as confirmed in Tab.~\ref{reb:table:panohead_comparison}. However, it cannot enable 360$^\circ$ rendering.

% The use of PanoHead-inverted pseudo-multiview data introduces a strong extra supervision signal, complicating fair comparison.

% \noindent\textbf{Cross-identity Reenactment.}

% \subsubsection{Additional Ablation Study}
% \noindent\textbf{Local Window in Cross Attention.}

% \begin{table*}[tb]
%   \centering
%   \resizebox{1\linewidth}{!}{
%   \begin{tabular}{l|ccccccc|ccccccc} 
%   \toprule[1.2pt]
%   \multirow{2}{*}{Method}  &  \multicolumn{7}{c|}{HDTF}  &  \multicolumn{7}{c}{MEAD} \\
%              & PSNR $\uparrow$  & SSIM $\uparrow$   & LPIPS $\downarrow$ & CSIM $\uparrow$  & AKD $\downarrow$ & AED $\downarrow$ & APD $\downarrow$
%              & PSNR $\uparrow$  & SSIM $\uparrow$   & LPIPS $\downarrow$ & CSIM $\uparrow$  & AKD $\downarrow$ & AED $\downarrow$ & APD $\downarrow$
%              \\
%     \midrule
%     w/o local window
%     & 26.57 & 0.8643 & 0.1896 & 0.8607 & 2.302 & 0.3387 & 0.5359
%     & 17.12 & 0.7462 & 0.3435 & 0.6635 & 5.195 & 0.4920 & 2.3008
%     \\
%     \midrule
%     Ours 
%     & \textbf{26.61} & {0.8642}  & {0.1900}  & 0.8622  & \textbf{2.287}  & \textbf{0.3318}   & \textbf{0.5286} 
%     & \textbf{17.20} & {0.7495} & \textbf{0.3402} & \textbf{0.6765} & 4.844 & 0.5037 & \textbf{1.9484}
%     \\ 

%     \bottomrule[1.2pt]
%   \end{tabular}
%   }
%   % \vspace{-8pt}
%   \caption{Ablation study on the local windows in cross attention. We present the results for self-reenactment on the HDTF~\cite{zhang2021flow} and multi-view MEAD~\cite{wang2020mead} datasets. 
%   }
%   \label{table:ablation_window}

%   % \vspace{-6pt}
% \end{table*}

\section{Limitations}

\begin{figure}[tb]
    \centering
   \includegraphics[width=1\linewidth]{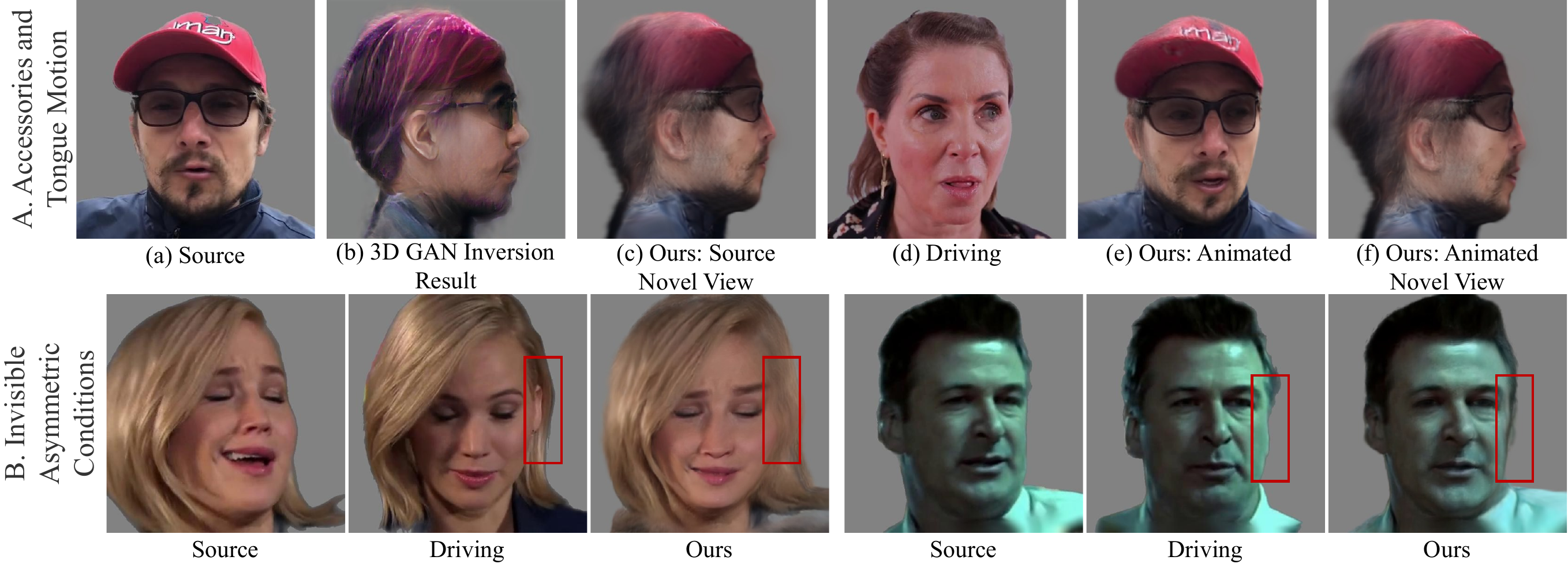}
  \vspace{-18pt}
  \caption{Failure cases of our method. It fails to reconstruct accessories faithfully and cannot transfer motions that FLAME cannot model, such as the tongue movement. It can also fail when the asymmetric conditions in the source image are occluded.
  }

  \label{fig:failure}
  % \vspace{-15pt}
\end{figure}

Despite the capability of our method to generate animatable 360$^\circ$ avatars, it still has several limitations that can be addressed in the future. First, our method cannot reconstruct accessories well, such as hats and glasses. The reason is that our method highly depends on the pretrained 3D GAN and its inversion for global full-head feature extraction and novel view supervision, and the inversion method~\cite{bilecen2024dual} is constrained by the latent space of the pretrained 3D GAN~\cite{an2023panohead}, which leads to noticeable artifacts when handling out-of-domain accessories, as shown in Fig.~\ref{fig:failure} A(b) and A(c). Besides, we obtain local UV feature maps by projecting points on the refined FLAME mesh to the 2D feature space, and the refined FLAME mesh cannot model glasses well, which makes our method treat glasses as textures on face surfaces, as shown in Fig.~\ref{fig:failure} A(c). Second, our performance on motion transfer is constrained by the estimated FLAME motion parameters. The limitation mainly comes from errors in the estimated parameters and the inability of FLAME to model certain motions, such as the tongue movement, as shown in Fig.~\ref{fig:failure} A(e) and A(f). \textcolor{black}{Third, our symmetric assumption can fail if the asymmetric conditions in the source image are hidden, such as the ear and lighting in Fig.~\ref{fig:failure} B.}
 
% haven't model expression-related wrinkles...

%%%%%%%%%%%%%%%%%%%%%%%%%%%%%%%%%%%%%%%%%%%%%%%%%%%%%%%%%%%%

\end{document}